\documentclass[review]{elsarticle}

\usepackage{hyperref}

\journal{Computing Research Repository}

\usepackage{url, amssymb, amsmath}
\usepackage{caption}
\usepackage{subcaption}
\usepackage{multirow}
\usepackage{color}
\usepackage{soul}

\usepackage{listings}
\usepackage{bm}
\usepackage[normalem]{ulem}
\DeclareMathOperator*{\argmax}{argmax}

\begin{document}

\begin{frontmatter}

\title{Explain and Conquer: Personalised Text-based Reviews to Achieve Transparency}



\author[mainaddress]{Iñigo~López-Riobóo~Botana\corref{mycorrespondingauthor}}
\ead{inigo.lopezrioboo.botana@udc.es}
\author[mainaddress]{Verónica~Bolón-Canedo}
\ead{veronica.bolon@udc.es}
\author[mainaddress]{Bertha~Guijarro-Berdiñas}
\ead{berta.guijarro@udc.es}
\author[mainaddress]{Amparo~Alonso-Betanzos}
\ead{ciamparo@udc.es}

\cortext[mycorrespondingauthor]{Corresponding author}
\address[mainaddress]{Research Center on Information and Communication Technologies (CITIC) - Universidade da Coruña. Campus de Elviña, 15071 A Coruña, España.}

\begin{abstract}

There are many contexts in which dyadic data are present. Social networks are a well-known example. In these contexts, pairs of elements are linked building a network that reflects interactions. Explaining why these relationships are established is essential to obtain transparency, an increasingly important notion. These explanations are often presented using text, thanks to the spread of the natural language understanding tasks.

Our aim is to represent and explain pairs established by any agent (e.g., a recommender system or a paid promotion mechanism), so that text-based personalisation is taken into account. We have focused on the TripAdvisor platform, considering the applicability to other dyadic data contexts. The items are a subset of users and restaurants and the interactions the reviews posted by these users. We propose the PTER (Personalised TExt-based Reviews) model. We predict, from the available reviews for a given restaurant, those that fit to the specific user interactions.

PTER leverages the BERT (Bidirectional Encoders Representations from Transformers) transformer-encoder model. We customised a deep neural network following the feature-based approach, presenting a LTR (Learning To Rank) downstream task. We carried out several comparisons of our proposal with a random baseline and other models of the state of the art, following the EXTRA (EXplanaTion RAnking) benchmark. Our method outperforms other collaborative filtering proposals.

\end{abstract}


\end{frontmatter}

\section{Introduction}
\label{sec:introduction}

Dyadic data refer to a domain with two finite sets of elements in which observations are made for dyads (i.e., pairs with one element from each set). These pairs are connected considering $N-M$ relationships. That is, an item from the first set can be linked to $M$ different items from the second set and an item from the second set can be linked to $N$ different items from the first set. Let us assume that the two kind of objects correspond to a set of users $U$ and a set of items $I$. Data $D$ is organised in dyads (d) as follows:

\begin{align}
     D  = \{d_0, ..., d_{(n-1)}\} : d_i = (u_j, it_k, int_{jk}),~ u_j \in U, ~ it_k \in I, 
     \label{form:dyad_general}
\end{align}

\noindent where $int_{jk}$ is the interaction established between $u_j$ and $it_k$. 

\begin{figure}[!htb]
    \centering
    \includegraphics[width=0.6\textwidth]{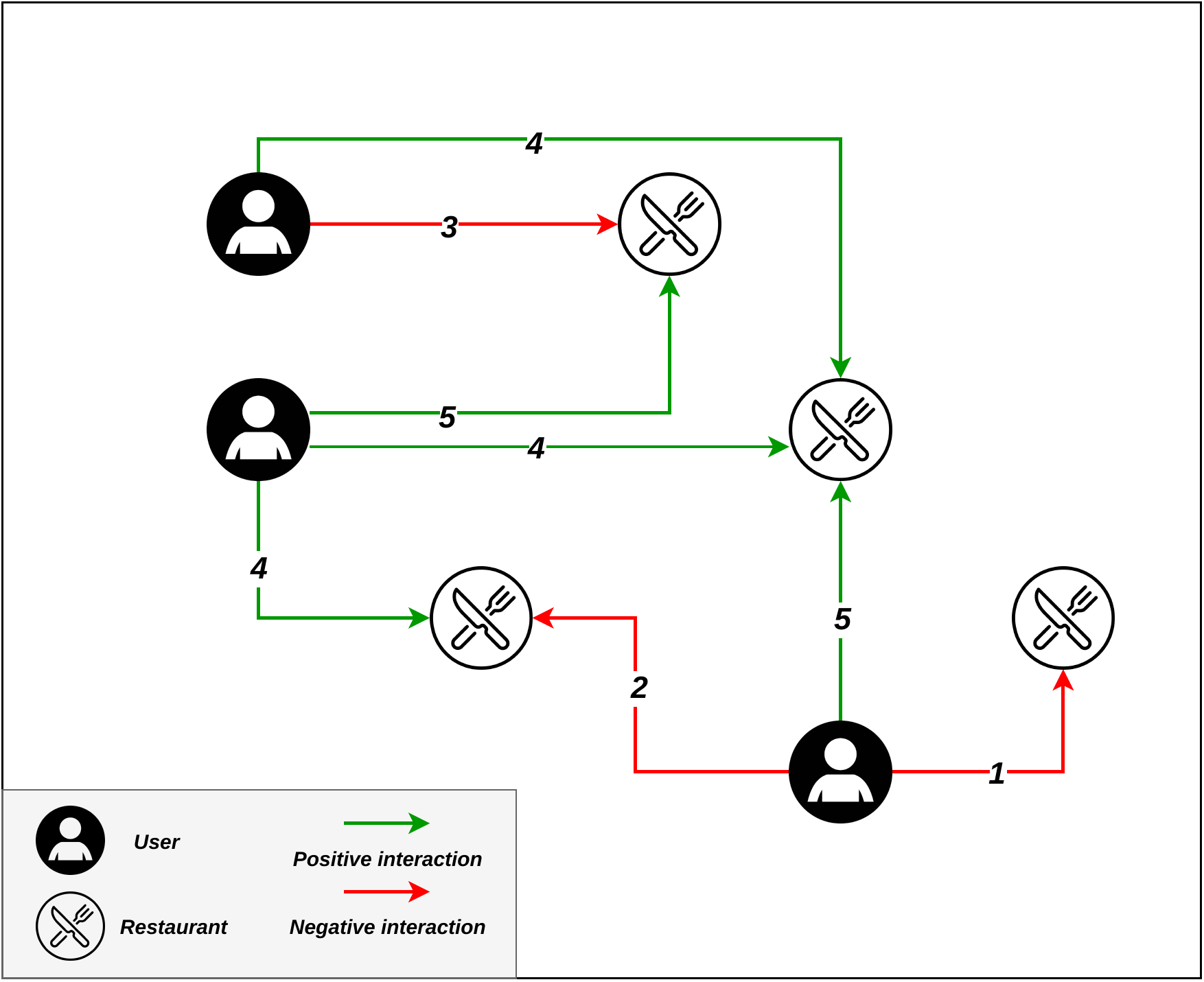}
    \caption{TripAdvisor interactions represented as a directed graph with labelled edges.}
    \label{fig:graph_approach}
\end{figure}

In the social network domain, interactions are established through different mechanisms (e.g., giving likes, uploading photos, posting comments and so on). Since we are exploring the transparency and explanation of the network in the NLU (Natural Language Understanding) domain, we have restricted these interactions to the text in the context of TripAdvisor (i.e., reviews). In this regard, we have adapted the formulation as follows:

\begin{align}
     D  = \{d_0, ..., d_{(n-1)}\} : d_i = (u_j, res_k, text_{jk}), ~u_j \in U, ~ res_k \in R, 
     \label{form:dyad_tripadvisor}
\end{align}

\noindent where $res_k$ is a restaurant from the set $R$ and $text_{jk}$ the review the user $u_j$ posted about the restaurant $res_k$. We can represent these interactions using a directed graph. These structures are represented by nodes connected using directed edges. These edges can be labelled, specifying weights. Adapting this idea to the TripAdvisor social network, as depicted in Figure \ref{fig:graph_approach}:

\begin{enumerate}
    \item \textit{Nodes}: represent the users $u_j \in U$ and restaurants $res_k \in R$. Users establish interactions with restaurants by posting reviews ($u_j \to res_k$) but not in reverse ($res_k \not\to u_j$).
    \item \textit{Edges}: represent the interactions between the nodes. Focusing on directed and text-based interactions, an edge ($u_j \to res_k$) defines a dyad ($u_j, res_k, text_{jk}) \in D$.
    \item \textit{User context}: we define the context $C(u_j)$ of a user $u_j$ as the set of interactions:
    \begin{equation}
        C(u_j) = \{(u_j \to res_k) ~ : ~ res_k \in R\}.
        \label{form:user_context_tripadvisor}
    \end{equation}
    We defined these contexts in a city-based scenario $C(u_j)_{city}$ to simplify and demarcate the graphs.
    \item \textit{Weights}: the interactions ($u_j \xrightarrow{s} res_k$) could be either positive or negative, according to the integer score $s \in [1,5]$ associated with the review. We defined 
    negative interactions as $s \in [1,3]$ and positives as $s \in [4,5]$.
\end{enumerate}

When the users interact (i.e., posting new comments) in TripAdvisor, they generate new pairs in the network, inherently explained with text since these pairs are supported by the authorship of the reviews. However, new pairs can also be suggested by other more opaque methods, such as recommender systems or paid promotion mechanisms. These methods leverage the available interactions and other features to match users $U$ with restaurants $R$. Despite the fact that these systems thoroughly proved their success, there is a gap in the research field related to the explanation of why these new pairs are suggested.

\begin{figure}[!htb]
    \centering
    \includegraphics[width=0.7\textwidth]{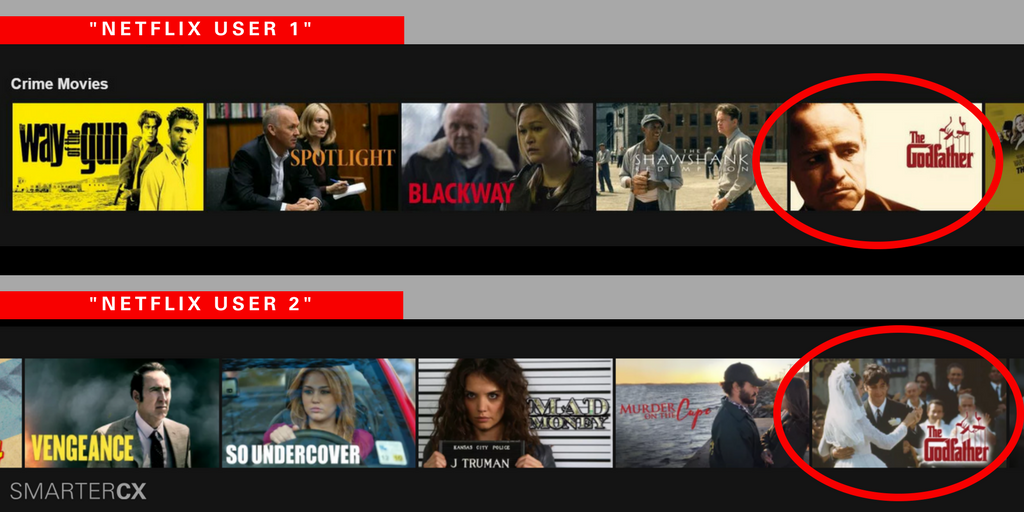}
    \caption{The same movie (The Godfather) is presented to several users. The thumbnails have noticeable differences. In the first row, the film is presented with a closeup view of a serious face (oriented to the thriller genre). However, in the second row the film is presented with a happy couple dancing during their wedding (related to sentiments like happiness or genres like comedy) \cite{netflix_thumbnails}.}
    \label{fig:thumbnails_netflix_1}
\end{figure}

From a general business point of view, recommender systems and paid promotion mechanism methods improve on customer satisfaction. In the case of recommender systems, the impact generating business value is noticeable. Netflix, for example, disclosed in a blog post \cite{blog_netflix_2017} that \textit{``(...) 75\% of what people watch is from some sort of recommendation (...)"} . Moreover, Netflix still grows its subscriber base by about 10\% yearly (i.e., 6 million new subscribers per year) and has identified over 2,000 ``taste communities" with specific content preferences \cite{blog_netflix_2021}. Furthermore, Netflix personalised recommendations have increased user engagement and helped to reduce customer churn over the last years \cite{netflix_recommender_system}. Netflix thumbnails are used to present items (films or series) to the users. The same item can be presented to the users in very different ways, depending on their interactions in the platform and their interests, so that these thumbnails change accordingly \cite{netflix_thumbnails}. This intuitive idea is depicted in Figure \ref{fig:thumbnails_netflix_1}.

Amazon, on its side, attributes up to 35\% of the revenue to the cross-selling system (i.e., recommendation related to the features ``Frequently bought together" and ``Customers who bought this item also bought") \cite{forbes_amazon_analytics}. Recommendations are reinforced if we look at the most well-known online platforms, where a major part of the home page and interface is used to show personalised content to the users. Personalisation really matters to customers, and to business. 93\% of companies with advanced recommendation strategies in their processes have significantly increased their revenue in recent years \cite{recommendation_revenue_study_2019}. The later report \cite{recommendation_revenue_study_2019} states that 77\% of businesses that exceeded their revenue goals in 2019 have a documented personalisation strategy, while 74\% have a dedicated budget for it.

Thus, personalisation and explanation play a crucial part in the acceptability of the user. Particularly in the context of TripAdvisor, it is not only important to present an interesting restaurant for the user, but also it is of utmost importance how this presentation is made. Our intention is to fill this gap by proposing a general framework for dyadic data text-based explanation, with a practical example in this social network. When one of the aforementioned methods (recommender system or paid promotion mechanism) matches a user from $U$ to a restaurant from $R$, our model PTER (Personalised TExt-based Reviews) will choose for that user, and from among the reviews available in the system, the text that he/she would have written as a review if he/she had visited the restaurant before. Therefore, our selected textual representation should resemble the authorship of the reviews, acting as a personalised summary of why that restaurant was linked to the user. There are more probabilities for the user to visit the restaurant if the algorithm shows a text that identifies his/her preferences. In other words, provided that we know in advance the new suggested pairs by these systems (e.g., recommender systems, paid promotion mechanisms and so on), we aim to answer the question ``\textit{how to present and explain it?}". Our contributions are as follows:

\begin{enumerate}
    \item We leveraged the deep transformer-encoder model BERT (Bidirectional Encoders Representations from Transformers) \cite{bert} as a pre-trained surrogate model to provide text-based explanations in a LTR (Learning To Rank) text retrieval task. To the best of our knowledge, this has not been explored in the TripAdvisor platform. We propose a way to explain dyads using the available reviews of the restaurants.
    \item We present a new data labelling proposal in dyadic environments for supervised NLP downstream tasks. We provide experimental results proving the consistency of our ground truth definition. This problem definition regarding explanation in dyadic data contexts can be further explored and proposed as a new benchmark task in NLU.
    \item We designed our own evaluation framework based on a clustering process to assess the human suitability of the explanations using a random adversary and an authorship reference, showing the significant improvement of our proposal. Moreover, we leveraged the EXTRA (EXplanaTion RAnking) benchmark datasets \cite{extra_datasets_github} to compare PTER with other state-of-the-art baselines.
    \item We publicly share six new TripAdvisor datasets \cite{tripadvisor_datasets} corresponding to the cities of Barcelona, Madrid, New York, New Delhi, London and Paris to support further research regarding dyadic data explanation tasks.
\end{enumerate}

The remainder of the article is organised as follows: Section \ref{sec:related_work} presents the state-of-the-art works in relation to the explanation of pairs (user, item) in recommender systems and particularly in the hospitality context. In Section \ref{sec:materials_methods}, we describe the data, the natural language pre-processing methods and the BERT model used in this work. In Section \ref{sec:proposed_method}, we explain our approach, defining the problem at hand and the topology of the PTER model, as well as the basis of our evaluation framework. In Section \ref{sec:experimental_results}, we continue with the experimental results. Finally, in Section \ref{sec:conclusions}, we summarise the conclusions and future research.

\section{Related Work}
\label{sec:related_work}

The most widespread methods to suggest new dyads in general networks are recommender systems. Learning to explain these recommendations is a hard task to do. The difficulty to generate and assess the explanations from the human perspective as well as the definition of standard evaluation frameworks has led to a myriad of approaches regarding explanation in this context. However, existing explainable recommender systems usually consider explanation as a side output of the recommendation model. Such models do not address explainability as the main task \cite{extra-framework}.

In these works, some approaches leverage deep neural networks and propose both feature extraction and explanation as a text retrieval task, using the available reviews written by the users. The DeepCoNN (Deep Cooperative Neural Network) \cite{explanations_retrieve_text_based_deepconn} consists of two parallel CNN (Convolutional Neural Networks) coupled in the last layers. One of the networks focuses on learning user behaviours exploiting reviews written by them, whereas the other one learns item properties from the reviews written for the item. Although this model represents both users and items in a joint manner using reviews, the text retrieval task is centred only on the feature extraction rather than on the explanation. TransNets \cite{explanations_retrieve_text_based_transnet} is an extension of the previous DeepCoNN model, which introduces an extra latent layer representing the user-target item pair. This model introduces the situation where the item is presented to the user before he/she have experienced it (i.e., no real review is available for that pair). Thus, it predicts an approximate and personalised representation of the target review, but only for feature extraction in rating prediction tasks. NARRE (Neural Attentional Regression model with Review-level Explanations) \cite{explanations_retrieve_text_based_narre} is another two parallel neural networks model based on attention mechanisms which describes both users and items in latent representations. It predicts precise ratings, learning the usefulness of each review. The highly-useful reviews obtained from the attention scores are presented as the review-level explanations. However, these partial explanations are side outputs of NARRE. Moreover, it is based on CNN transformations to extract the semantic features, rather than using state-of-the-art transformers. DER (Dynamic Explainable Recommender) \cite{explanations_retrieve_text_based_der} is a similar sentence-level CNN approach with attention mechanisms that takes into account the dynamic behaviour of user preferences over time, accounting for a time-aware GRU (Gated Recurrent Unit). It predicts ratings, providing adaptive recommendation explanations according to the user dynamic preference. Again, it has the same problems as NARRE. NERAR (Neural Explainable Recommender model based on Attributes and Reviews) \cite{nerar_model} combines the processing of item and user attributes (using a tree-based model) with the semantic features of reviews. This model uses a time-aware GRU to handle user preferences over time and CNN transformations to extract the semantic features of text. Again, it relies on a side output depending on the attention scores to provide partial explanations and it does not leverage state-of-the-art transformers to handle reviews. Interestingly, there are attempts \cite{extra-framework} to achieve standard evaluation of explainability for recommendations, following a ranking-oriented task.

A more different set of approaches focus on the text generation task to explain \cite{text_generation_caml, text_generation_nete, text_generation_nrt} or use pre-built templates \cite{text_templates_deaml, text_templates_caesar, text_templates_efm}. Other methods use visual data proposing image segmentation as part of the explanation to the recommendation \cite{explanation_image_segments, explanation_image_segments_2, explanation_image_segments_3}. Finally, textual counterfactual explanations for neural-based recommender systems are also explored in this field \cite{counterfactual_explanations_neural_recommenders}.

To the best of our knowledge, most approaches in the hospitality sector only focus on rating prediction tasks, sometimes considering side outputs which partially attain explainability. However, the idea of review-level explanation in a text retrieval task has not been widely explored in social networks yet. We propose a LTR task to explain dyadic data considering, but not limited to, the hospitality context of TripAdvisor.

\section{Data Preparation and Methods}
\label{sec:materials_methods}

In this section we will describe the NLP data preparation techniques and methods that have been used to carry out our work.

\subsection{Datasets}
\label{subsec:datasets}

In order to obtain the datasets to work with, we sought for common cities visited around the world. Our intention was to discover cities with a large amount of reviews written in English, since at this time we have defined the task in a monolingual approach. We finally chose Barcelona (1,620,343 city inhabitants), London (8,961,989 city inhabitants), Madrid (3,223,334 city inhabitants), New Delhi (257,803 city inhabitants), New York (8,175,133 city inhabitants) and Paris (2,175,601 city inhabitants). These datasets \cite{tripadvisor_datasets} were retrieved with our own ad hoc web scraper. Small datasets corresponding to medium to small cities were discarded due to scarcity of user interactions. Table \ref{tab:dataset_features_summary} summarises the features extracted from TripAdvisor. It is worth mentioning that we retrieved a representative amount of the reviews for each city, not the whole set of them.

\begin{table}[!htb]
\centering
\resizebox{\textwidth}{!}{%
\begin{tabular}{|c|c|c|}
\hline
\textbf{Feature name} & \textbf{Feature type} & \textbf{Description} \\ \hline
\textit{parse\_count} & Numerical (integer) & Corresponding no. of extracted review (auto-incremental) \\ \hline
\textit{user\_id} & Categorical (string) & Univocal ``internal" (UID\_XXXXXXXXXX) identifier of the user \\ \hline
\textit{author} & Categorical (string) & Univocal ``external" (@user) identifier of the user \\ \hline
\textit{restaurant\_name} & Categorical (string) & Name of the restaurant matching the review \\ \hline
\textit{rating\_review} & Numerical (integer) & Review score in the range {[}1-5{]} \\ \hline
\textit{sample} & Categorical (string) & ``Positive" score {[}4-5{]} or ``Negative" score {[}1-3{]} sample \\ \hline
\textit{review\_id} & Categorical (string) & Univocal ``internal" (review\_XXXXXXXXX) identifier  of the review \\ \hline
\textit{title\_review} & Text & Review title \\ \hline
\textit{review\_preview} & Text & Summarised review as a preview \\ \hline
\textit{review\_full} & Text & Complete review \\ \hline
\textit{url\_review} & Text & Url of the gathered review from TripAdvisor \\ \hline
\textit{date} & Date & Publication date in format (day, month, year) \\ \hline
\textit{city} & Categorical (string) & City of the restaurant which the review was written for \\ \hline
\textit{url\_restaurant} & Text & Restaurant url \\ \hline
\end{tabular}%
}
\caption{Features with data types and their descriptions.}
\label{tab:dataset_features_summary}
\end{table}

\begin{figure}[!htb]
    \centering
    \includegraphics[width=0.7\textwidth]{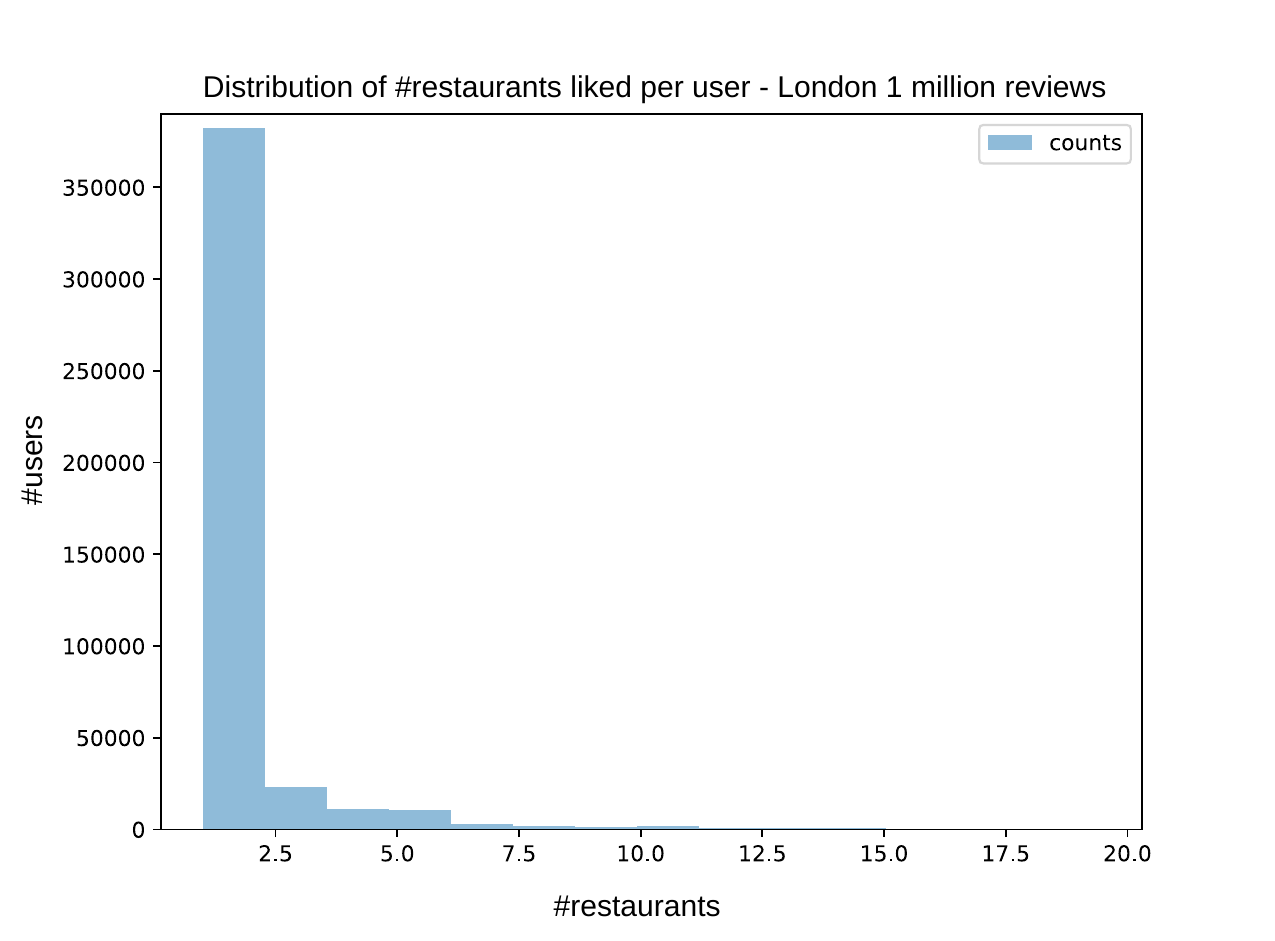}
    \caption{Distribution of \#restaurants liked per user (London dataset). Most of the distribution is concentrated in the first interval of the histogram (between 1 and 2 positive interactions per user), thus showing that there are a large number of inactive accounts in the TripAdvisor platform.}
    \label{fig:n_interactions_london}
\end{figure}

Some properties about extracted datasets are shown in Table \ref{tab:datasets_distribution_reviews_users}. Looking at that information, we concluded some insights about the users' behaviour and the quality of the data. A first observation is that users mostly tend to write positive reviews about restaurants. This situation is due to the \textit{Customer Complaint Iceberg} \cite{complaint_iceberg}. Good experiences encourage people to share information whereas negative experiences do not produce the same effect. In this way, the negative context of the restaurants is incomplete. A second observation is that the relation between positive reviews and the number of different users in the datasets follows a two-one rule. This means that for every two different reviews, there is a new user (or account) in the platform. This condition roughly fulfils for almost every dataset from Table \ref{tab:datasets_distribution_reviews_users}, checking the ``review/user ratio" column. It is worth mentioning that New Delhi has the highest ratio (2.06) whereas Madrid has the lowest (1.64). This will be determining in the experimentation Section \ref{sec:experimental_results}. Moreover, the distribution of restaurants liked per user shows that there are a lot of inactive accounts (i.e, with few reviews written), so their interactions are sparse (i.e., less ``dense"). Active users are uncommon, as depicted in the example of Figure \ref{fig:n_interactions_london}, where it can be observed that most of the distribution is concentrated in the first interval of the histogram (i.e., between 1 and 2 positive interactions per user). This is a core issue to address in the PTER model definition and in the dataset labelling process, since the learning result specially depends on the quality (``density" of user interactions) of input data.

\begin{table}[!htb]
\centering
\resizebox{\textwidth}{!}{%
\begin{tabular}{|c|c|c|c|c|c|c|c|}
\hline
\textbf{Dataset} & \textbf{Size} & \textbf{\begin{tabular}[c]{@{}c@{}}\#Reviews \\ (English)\end{tabular}} & \textbf{\begin{tabular}[c]{@{}c@{}}\#Positive \\ reviews \\ (English)\end{tabular}} & \textbf{\begin{tabular}[c]{@{}c@{}}\#Negative \\ reviews \\ (English)\end{tabular}} & \textbf{\begin{tabular}[c]{@{}c@{}}\#Different \\ users\end{tabular}} & \textbf{\begin{tabular}[c]{@{}c@{}}Review/user \\ ratio \end{tabular}} &
\textbf{\begin{tabular}[c]{@{}c@{}}\#Different \\ restaurants\end{tabular}} \\ \hline
New Delhi & 162.5 MB & 148,303 & 123,095 & 25,208 & 59,796 & 2.06 & 5,147 \\ \hline
New York & 559.1 MB & 517,604 & 424,823 & 92,781 & 218,738 & 1.94 & 1,715 \\ \hline
Madrid & 198.9 MB & 177,353 & 145,177 & 32,176 & 88,560 & 1.64 & 5,481 \\ \hline
London & 1.1 GB & 998,939 & 833,453 & 165,485 & 441,821 & 1.89 & 1,827 \\ \hline
Barcelona & 428.3 MB & 417,240 & 339,385 & 77,855 & 203,514 & 1.67 & 6,319 \\ \hline
Paris & 558.8 MB & 510,084 & 401,589 & 108,495 & 219,340 & 1.83 & 11,004 \\ \hline
\end{tabular}%
}
\caption{Distribution of reviews, restaurants and users. The ``review/user ratio" is computed comparing the ``\#Positive reviews" column with respect to the ``\#Different users" column. ``\#Different users" and ``\#Different restaurants" are computed considering the positive reviews written in English (i.e., with a score in the range $[4,5]$).}
\label{tab:datasets_distribution_reviews_users}
\end{table}

For the purpose of the task proposed in Section \ref{sec:introduction}, the datasets were arranged in dyads as formally described in Equation \ref{form:dyad_tripadvisor}. We only considered the positive interactions of the datasets due to (1) the scarcity of negative interactions as seen in Table \ref{tab:datasets_distribution_reviews_users} and (2) the scope of the task, centred on the text-based explanation of the new dyads $(u_j, res_k, text_{jk})$ suggested by the recommender system, which we assume to be positive relationships. Giving an example, if a recommender system (or other mechanism) presents to the user $u_j \in U$ a restaurant $res_k \in R$, it is expected that the user will like it. Therefore, the explanation for that suggested interaction, $text_{jk}$, should be in a positive direction. Interactions ($u_j \xrightarrow{s} res_k$) are thus restricted to $s \in [4, 5]$.

In order to solve this explainability task in dyadic environments, we followed a supervised LTR task. To do so, we required labelling the datasets previously arranged in dyads. Our aim was to represent the user contexts (i.e., set of interactions) as the ground truth definition of our approach, thus learning a function which maps every text to the aforementioned user contexts.

\subsection{Datasets labelling proposal}
\label{subsec:labelling_proposal}

For the sake of completeness and in order to mitigate the scarcity of interactions, we redefined the user context $C(u_j)$ from Equation \ref{form:user_context_tripadvisor} from previous Section \ref{sec:introduction} with a positive $(+)$ expansion $(*)$, as follows:
\begin{align}
    C^+*(u_j) & = C^+(u_j) \cup C^+_{u_j}(res_k), ~ \forall res_k \in R ~ : ~ \exists (u_j \xrightarrow{s} res_k),
    \label{form:expansion_user_context}
\end{align}

\noindent where $C^+(u_j)$ is the positive user context $C(u_j)$ from Equation \ref{form:user_context_tripadvisor} using only relations with $s \in [4, 5]$. $C^+_{u_j}(res_k)$ corresponds to the positive $(+)$ restaurant context of $res_k \in R$ considering user $u_j$, defined as:
\begin{equation}
    C^+_{u_j}(res_k) = \{(u_i \xrightarrow{s} res_k) ~ \forall u_i \in U : i \neq j, ~\exists (u_j \xrightarrow{s} res_k)\},
    \label{form:indegree_positive_res}
\end{equation}

\noindent which represents the rest of positive interactions with $s \in [4, 5]$ that $res_k$ has with different users $u_i \in U$ in addition to $u_j$. In the same way, the positive expansion of a user context $C^+*(u_j)$ can be demarcated in a city as $C^+*(u_j)_{city}$. 

Since we defined our downstream LTR task for dyadic data explanation as a supervised process, we required labelling our data. Our ground truth definition matches the aforementioned positive $(+)$ expansion $(*)$ of the user contexts given a city. Therefore, targets were built as the set:
\begin{equation}
    targets = \{C^+*(u_j)_{city} ~ \forall city \in C, ~ \forall u_j \in U^- : U^-\subset U\},
    \label{form:targets_equation}
\end{equation}
where $U^-$ is the subset of the most active users (i.e., with the highest amount of positive interactions, depending on a hyper-parameter later described) and $C$ the set of the six aforementioned cities.

Therefore, in every training sample the input corresponds to a single review posted about a restaurant $res_k \in R$ by any user $u_j \in U$. To encode targets (outputs),  we restrict to the subset $U^-$ and represent those targets with a binary vector of length $|U^-|$ in which each position $j$ corresponds to user $u_j$. The step-by-step labelling process, depicted in Figure \ref{fig:multi_label_ground_truth_process}, is decomposed as follows: 

\begin{figure}[!htb]
    \centering
    \includegraphics[width=0.8\textwidth]{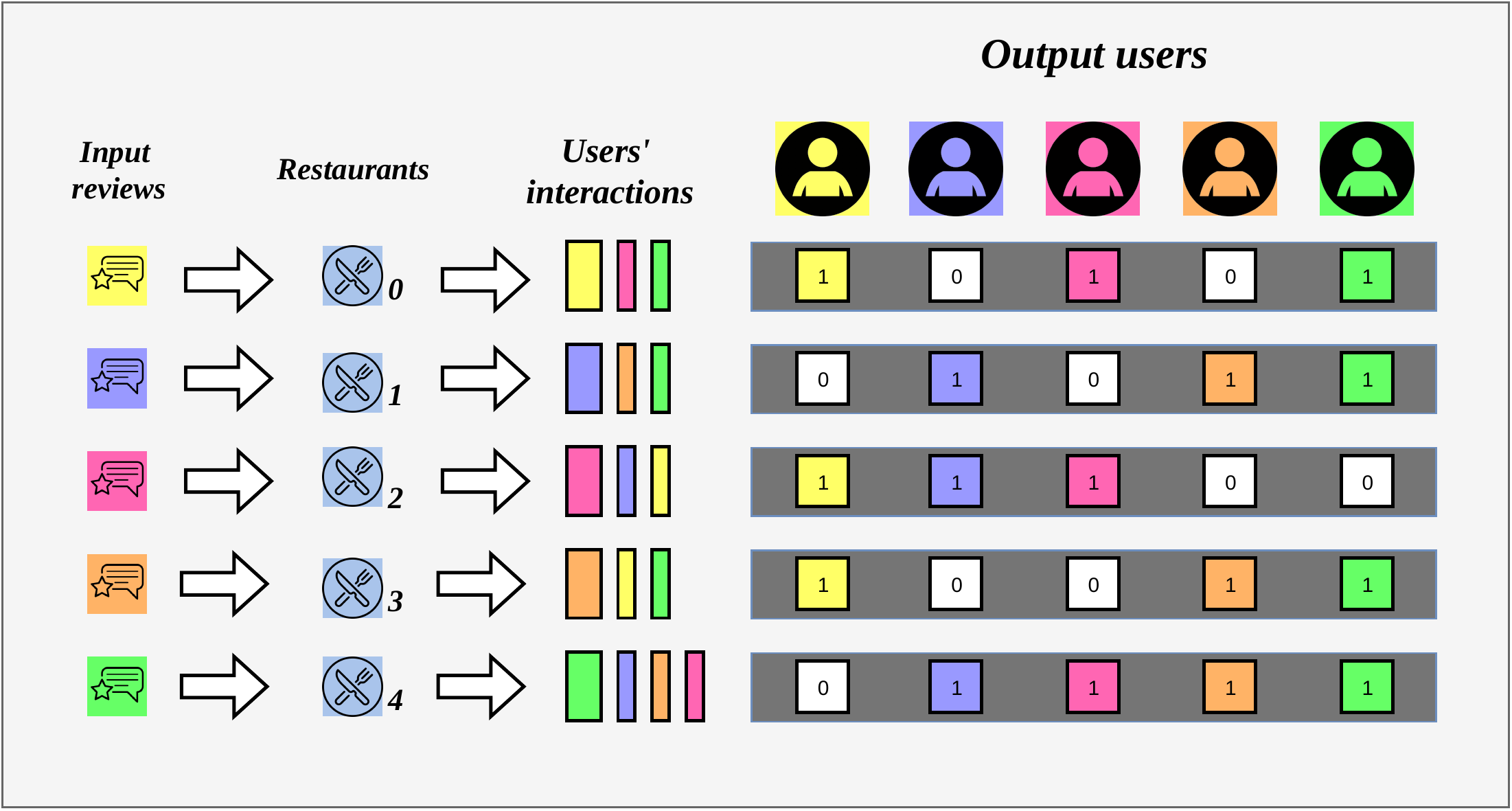}
    \caption{Each input is a review related to a restaurant. The output is set up as a binary matrix. Following the colour-based code, the diagonal of the matrix defines the set of interactions $C^+(u_j)\subset C^+*(u_j)$, whereas the other labelled positions correspond to the set of interactions $C^+_{u_j}(res_k)\subset C^+*(u_j)$. Zero-valued positions, in white colour, mean that the user did not post that review and did not interact with that restaurant.}
    \label{fig:multi_label_ground_truth_process}
\end{figure}

\begin{itemize}
    \item A label $j$ is flagged with a 1 in the output vector if the user $u_j \in U^-$ is the author of the input review (authorship criterion). This corresponds to the set of interactions $C^+(u_j)$ in Equation \ref{form:expansion_user_context}.
    \item A label $j$ is flagged with a 1 in the output vector if the input review belongs to a restaurant $res_k \in R$ that the user $u_j \in U^-$ positively interacted with. This corresponds to the set of interactions $C^+_{u_j}(res_k)$ from Equation \ref{form:expansion_user_context}. The assumption of the user context expansion $(*)$ was carried out to mitigate the problem of scarcity in the number of interactions per user account, as seen in Figure \ref{fig:n_interactions_london} and Table \ref{tab:datasets_distribution_reviews_users}.
    \item A label $j$ is flagged with a 0 in the output vector if user $u_j$ did not post the input review (authorship) nor positively interacted with that restaurant. If the whole set of labels are set to zero for a given input review, the sample is discarded from the dataset $D$.
\end{itemize}

\subsection{BERT transformer-encoder model}
\label{subsec:bert}

In order to work with text and use it as input for machine learning models, we need to establish a mapping between natural language and a numerical vector space. This is done with word embeddings \cite{survey_embeddings_chapter, survey_embeddings_article}.

Given a set of tokens, the embedding function maps them to a real-valued vector. Usually, we get a vector for each of the tokens, but we can also get a pooled representation of the sequence. The great advantage of this representation is the dimensionality reduction in comparison with the sparsity of the word space representations (i.e., the number of numerical features is much smaller than the size of the vocabulary). Word sequences that are closer in the vector space are expected to have similar meanings in the natural language domain.

BERT \cite{bert} is a deep transformer-encoder model capable of learning the bi-directional context of sequences \cite{bert_architecture_based}, referring to the transformer-encoder architecture. This model fits into numerous NLP tasks (e.g., text classification, sentiment analysis and so on), just adding a handful extra layers to the base pre-trained model. BERT is flexible and adaptable thanks to its input embedding definition \cite{bert}. We followed a feature-based approach of the model \cite{tuning_vs_feature_based}, where pre-trained weights of BERT are frozen. As a transfer learning approach, we used the concatenation of the last four hidden layers \cite{pooling_strategy_bert, 4_last_layer_strategy_bert} as the input features for the PTER model.

We leveraged the BERT-\textit{base-uncased} \cite{bert_implementation}, which has 12 layers (i.e., transformer blocks), a 768 hidden layer size (i.e., the length of the word embeddings) and 12 self-attention heads \cite{self-attention-bert}. The total number of parameters is 110 million. To address the input sequences, we followed the input embedding transformation using the tailored BERT-\textit{base} tokenizer \cite{bert_implementation}.

Besides, before this word embedding generation we made some NLP pre-processing steps with the reviews: (1) we discarded punctuation from the input, (2) transformed them to lower-case (uncased version of BERT), and finally (3) we made a head truncation policy of 512 tokens over the sequences, since it is the maximum input sequence length defined in BERT model. Other well-known pre-processing steps (such as stopwords, lemmatization) were omitted in this stage because they could degrade the bi-directional context of the reviews, taking into account the attention mechanism of BERT \cite{self-attention-bert}.

\section{Proposed Method}
\label{sec:proposed_method}

\begin{figure}[!htb]
    \centering
    \includegraphics[width=0.8\textwidth]{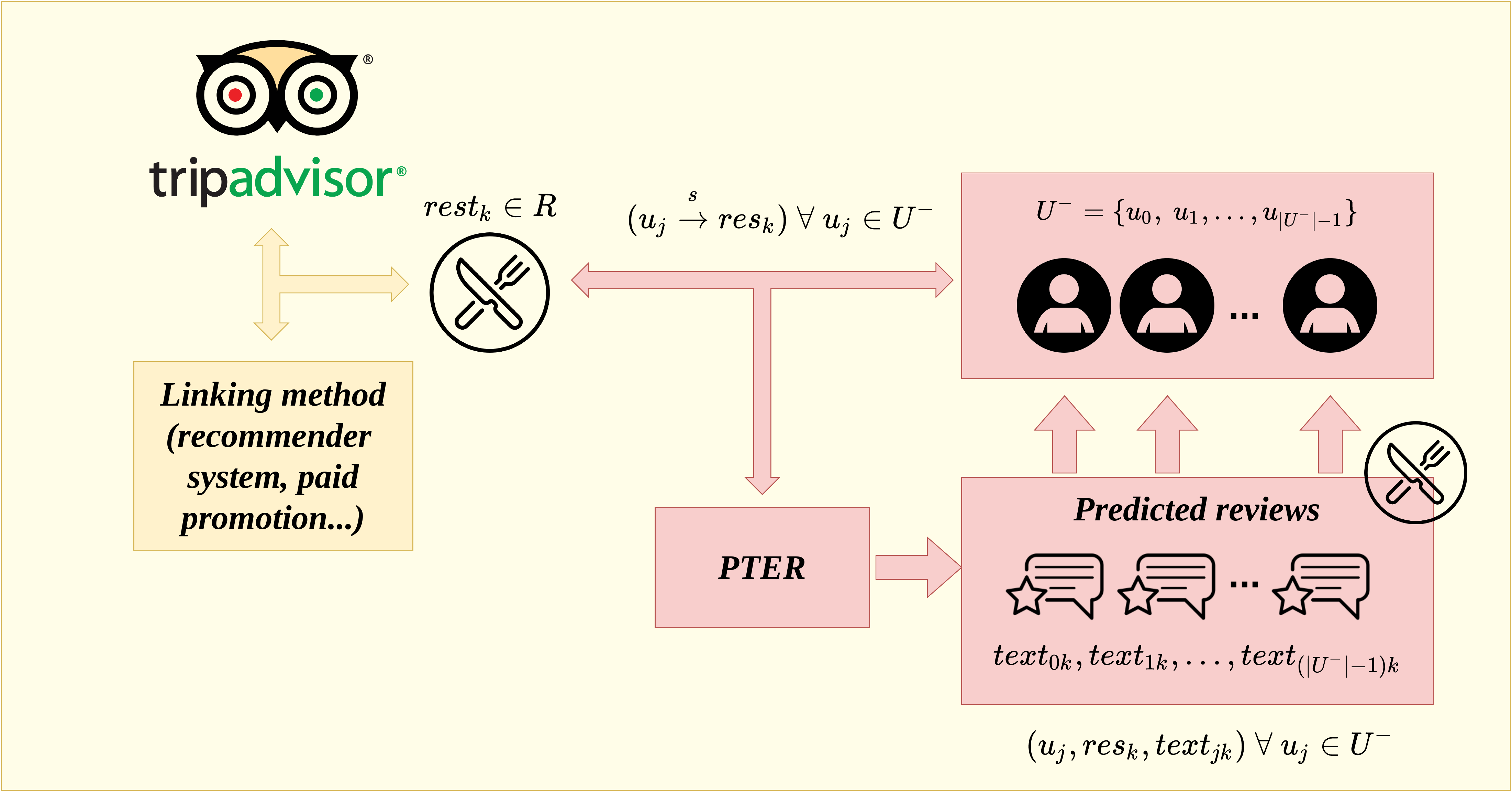}
    \caption{A linking method (e.g., a recommender system, a paid promotion strategy...) is responsible for suggesting new interactions ($u_j \xrightarrow{s} res_k$) assuming $s \in [4,5]$. Afterwards, the PTER model explains them, defining a new and explained dyad $(u_j, res_k, text_{jk})$.}
    \label{fig:overall_proposed_approach}
\end{figure}

Given the subset of users $U^- \subset U$ from a $city \in C$ and a restaurant $res_k \in R$ suggested to the users by any mechanism (e.g., a recommender system, a paid promotion method...), our model PTER will predict the ``best" review $text_{jk}$ for each user $u_j \in U^-$, defining a new and explained dyad $(u_j, res_k, text_{jk})$. These ``best" reviews explain, in a personalised way, the newly suggested interactions ($u_j \xrightarrow{s} res_k$) assuming $s \in [4,5]$, as depicted in Figure \ref{fig:overall_proposed_approach}.

As stated before, this is a LTR task. Considering the positive reviews of the restaurant $res_k \in R$, we aim at learning the following function:

\begin{equation}
f: \Big(r^+ \leadsto (C^+*(u_j)_{city})\Big)~\forall u_j \in U^-, \forall r^+ \in reviews^+(res_k) ,~ city \in C,
\label{form:general_f_learn}
\end{equation}

\noindent where $r^+$ is a positive review for restaurant $res_k$, $C^+*(u_j)_{city}$ is the positive $(+)$ expansion $(*)$ of the user context from Equation \ref{form:expansion_user_context}, and $reviews^+(res_k)$ is the set of positive reviews of that restaurant. Following the idea of the LTR task, we must deal with the probabilities $Pr(r^+|u_j)$ for a user $u_j \in U^-$ of liking each of the positive reviews $r^+ \in reviews^+(res_k)$. We aim at predicting the review $r^*$ whose probability value is the highest, which corresponds to the ``best" explanation to the new suggested interaction ($u_j \xrightarrow{s} res_k$):

\begin{equation}
    r^* = text_{jk} = \argmax_{r^+ \in reviews^+(res_k)} \Pr(r^+|u_j). 
    \label{form:max_prob_f_learn}
\end{equation}

We defined a \textit{multi-label classification problem to solve the supervised LTR task}, considering the ground truth definition from Subsection \ref{subsec:labelling_proposal}. In this way, \textit{each output label $j$, corresponding to the user $u_j \in U^-$, can be addressed independently}. The model can behave better for some users and worse for others, depending on the quality (and completeness) of their contexts $C^+*(u_j)_{city}$. 

\begin{figure}[!htb]
    \centering
    \includegraphics[width=0.9\textwidth]{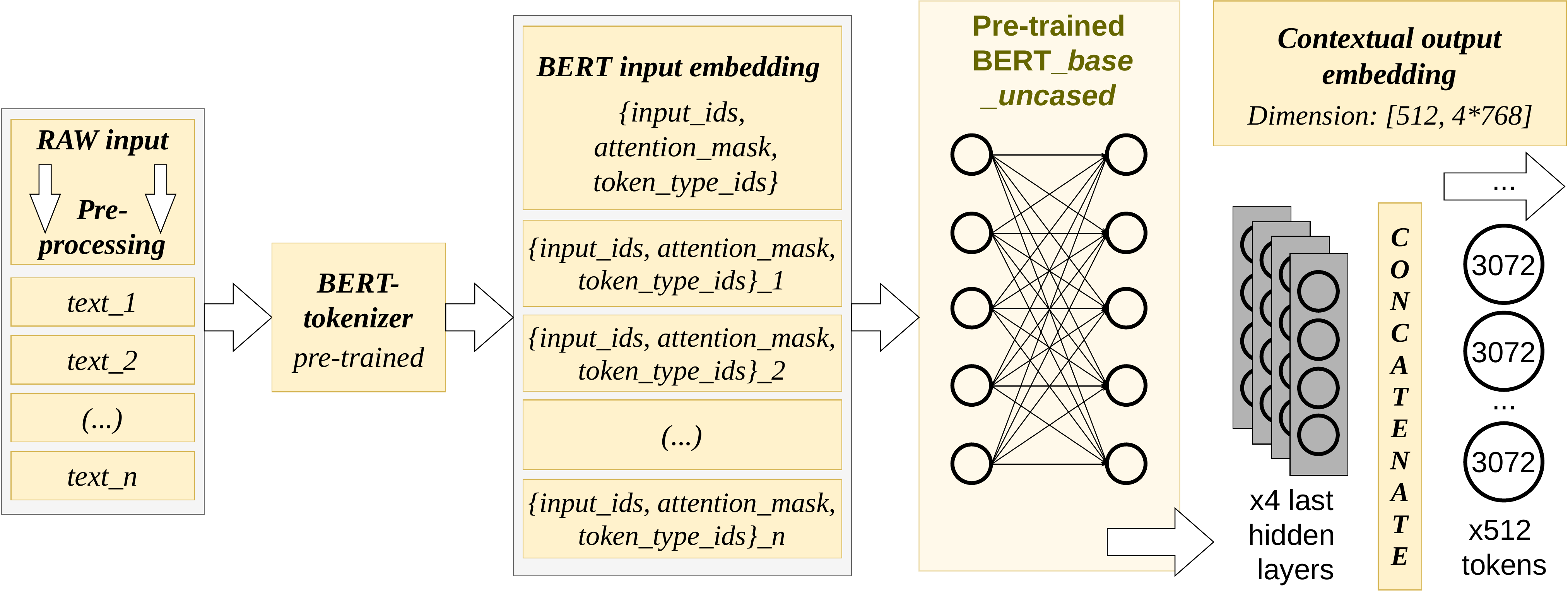}
    \caption{First block of the PTER model. Each raw input review is pre-processed and then translated to the BERT input embeddings using the BERT-\textit{tokenizer}. The contextual embedding is the concatenated representation of the hidden states of the last four hidden layers of BERT. These contextual embeddings are fed into the second block of our model.}
    \label{fig:network_architecture_block_1}
\end{figure}

\begin{figure}[!htb]
    \centering
    \includegraphics[width=0.9\textwidth]{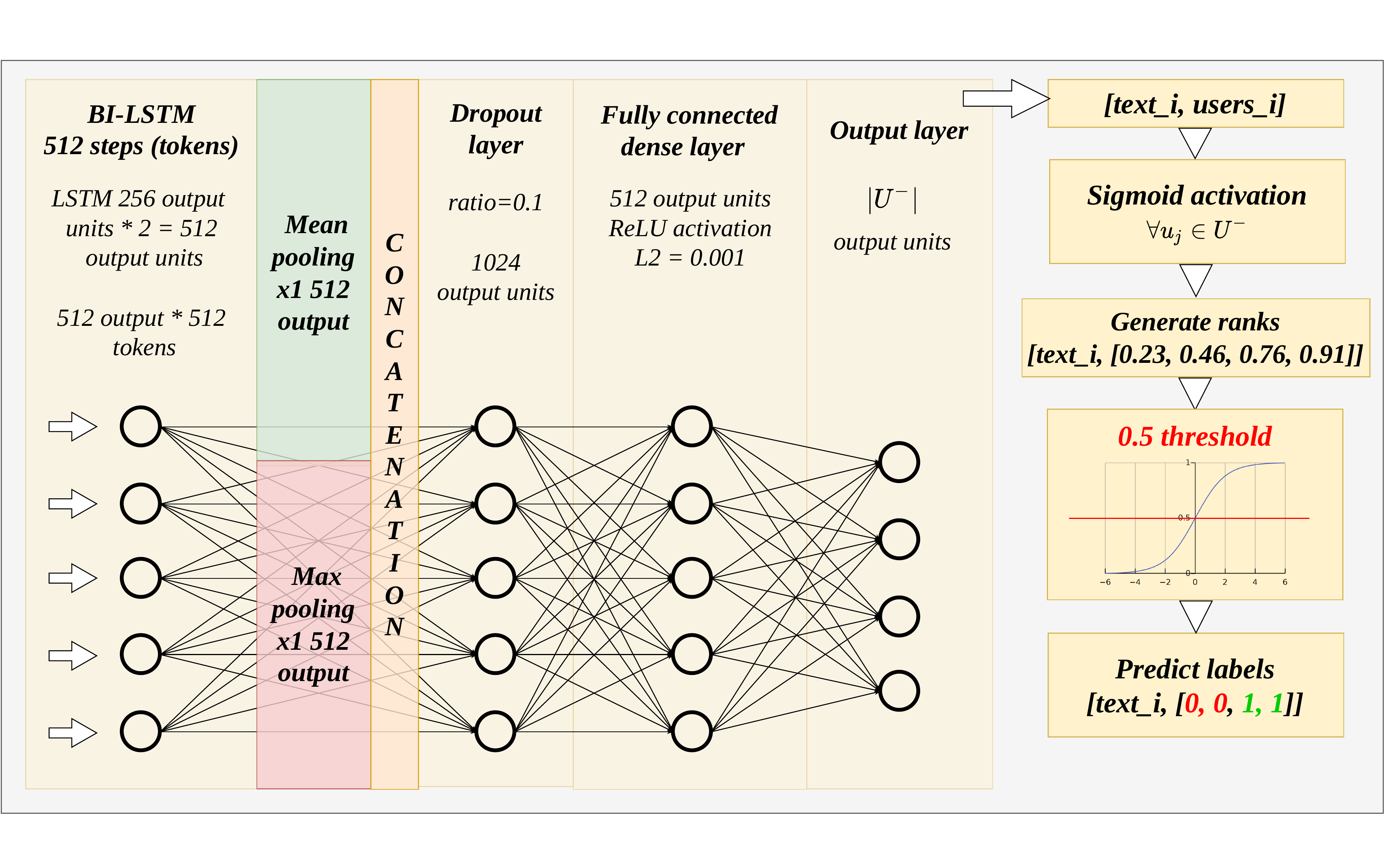}
    \caption{Second block of the PTER model, made up of the customised layers for the downstream LTR supervised task. The problem is addressed with multi-label classification. Inputs to this part of the model correspond to the contextual BERT embeddings.}
    \label{fig:network_architecture_block_2}
\end{figure}

As stated in Subsection \ref{subsec:bert}, PTER follows the \textit{feature-based} approach of the pre-trained BERT-\textit{base-uncased} model \cite{bert_implementation}. We focused on the hidden states of the last four layers to build what we called the \textit{contextual embedding}, following a transfer learning task. As a reminder, each sample in the training dataset has (1) the pre-processed input review (transformed to the input embeddings using the BERT-\textit{tokenizer}) and (2) the target binary vector. The PTER model has two neural network blocks, organised as follows:

\begin{enumerate}
    \item \textit{The mapping block}: it transforms natural language reviews to dense representations of real-valued vectors. This part of the model is straightforward, corresponding to the aforementioned BERT-\textit{base-uncased} model, depicted in Figure \ref{fig:network_architecture_block_1}. A concatenation layer links the hidden states of the last four hidden layers of the transformer, of size 768. This turns out in a $768 \times 4 = 3072$ \textit{contextual embedding} for each token in the input review (limited to a maximum of 512 tokens).
    
    \item \textit{PTER customised layers}: these layers tackle the multi-label classification problem for the supervised LTR task. Figure \ref{fig:network_architecture_block_2} depicts the overall design. First, the contextual embeddings are fed into a Bi-LSTM (Long Short-Term Memory) network \cite{rnn_and_lstm_explained} as time series data. The purpose of this layer is the classification of sequences, hence its suitability for NLP tasks. We present the contextual embeddings to the network as 512 periods or ``steps" to match with the maximum number of tokens allowed in the input sequences. The LSTM network was configured with 256 output units. The bi-directional approach of the LSTM learns the sequence in both the forward and backward directions, duplicating the output units. Moreover, the Bi-LSTM was configured to output the hidden states in every time step, generating a $256 \times 2 = 512$ output per input contextual embedding. After that, average and maximum pooling are addressed to reduce the output hidden states of every time step to a single output \cite{max_pooling_bilstm, generalized_pooling_bilstm} (i.e., summarising the time series data). A concatenation layer links both pooled representations, resulting in a 1024 dense representation of the whole input review. 
    
    Overfitting issues are faced dropping output units \cite{dropout_article_explained} with a dropout layer (with 0.1 ratio). Then, a fully connected dense layer with ReLU activation \cite{relu_function_explained} applies L2 weight regularisation (with 0.001 ratio) and decreases the dense representation size by half, with 512 output units. The output layer with size $|U^-|$ applies a sigmoid activation function that maps logits (i.e., $weights \times inputs + bias$) to real values in the range $(0, 1)$.

    PTER predicts the probabilities $\Pr(r^+|u_j)$ $\forall r^+ \in reviews^+(res_k)$ and for every user $u_j \in U^-$, given a restaurant $res_{k} \in R$. We used these probabilities to generate the rankings of reviews, following the idea of the LTR supervised task. Then, we computed the ``best" reviews, as stated in Equation \ref{form:max_prob_f_learn}. As we will see later, when describing the evaluation framework, the ``best" reviews take into account more than the ranking position. The standard threshold of 0.5 was used at the output of the sigmoid activation function to determine which labels are activated and which are not.
\end{enumerate}

In our approach, we faced the problem of sparsity in user interactions, stated in Subsection \ref{subsec:datasets}. We defined the previous concept of ``active users" from Equation \ref{form:targets_equation} as a model hyper-parameter. As a reminder, $U^- \subset U$ is the top $|U^-|$ users from a city context with the highest amount of interactions. It is worth mentioning that these top users determine the size of the output layer, as depicted in Figure \ref{fig:network_architecture_block_2}. The model will only consider them to make predictions and generate the rankings. This filtering criterion entails an important trade-off between the number of output users and the availability of training data:

\begin{itemize}
    \item \textit{If we reduce the number of output users}, there will be a larger amount of zero-valued target vectors that will be discarded from the training process, as already described. The advantage is that selected targets are less dispersed and easier to train with, but at the cost of losing training data and reducing the range of predictions. Moreover, filtering reviews in this way is making us discard some restaurants in each city.
    \item \textit{If we increase the number of output users}, we are increasing the scope of predictions, since increasing the number of users also entails an increase in the number of restaurants to be considered (see Figure \ref{fig:multi_label_ground_truth_process}). Therefore more data will be available to perform the task. However, selected targets are more dispersed and harder to train with.
\end{itemize}

\begin{figure}[!htb]
  \centering
  \begin{subfigure}[b]{0.6\textwidth}
    \includegraphics[width=\textwidth]{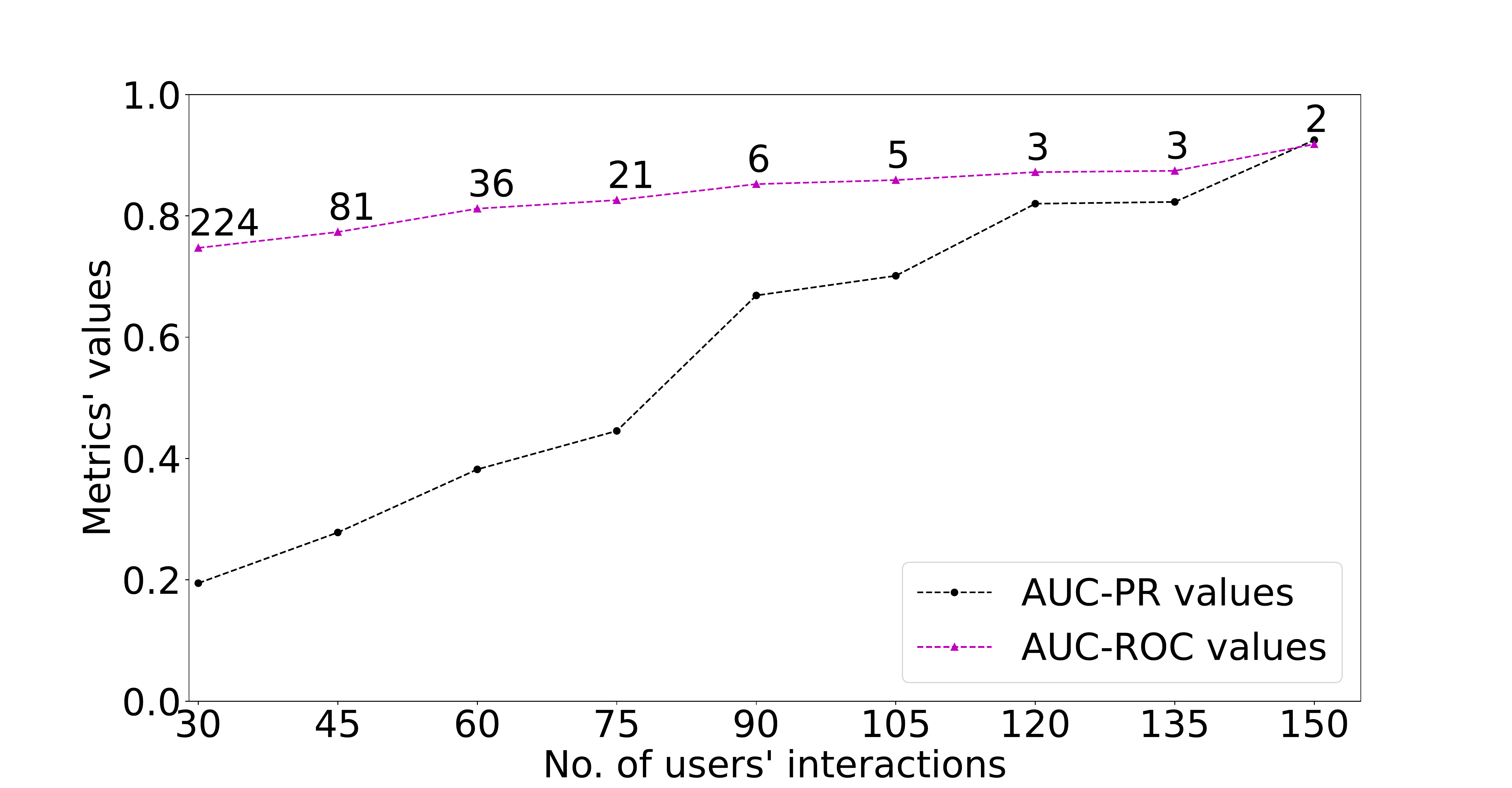}
  \end{subfigure}
  \par\bigskip
  \begin{subfigure}[b]{0.6\textwidth}
    \includegraphics[width=\textwidth]{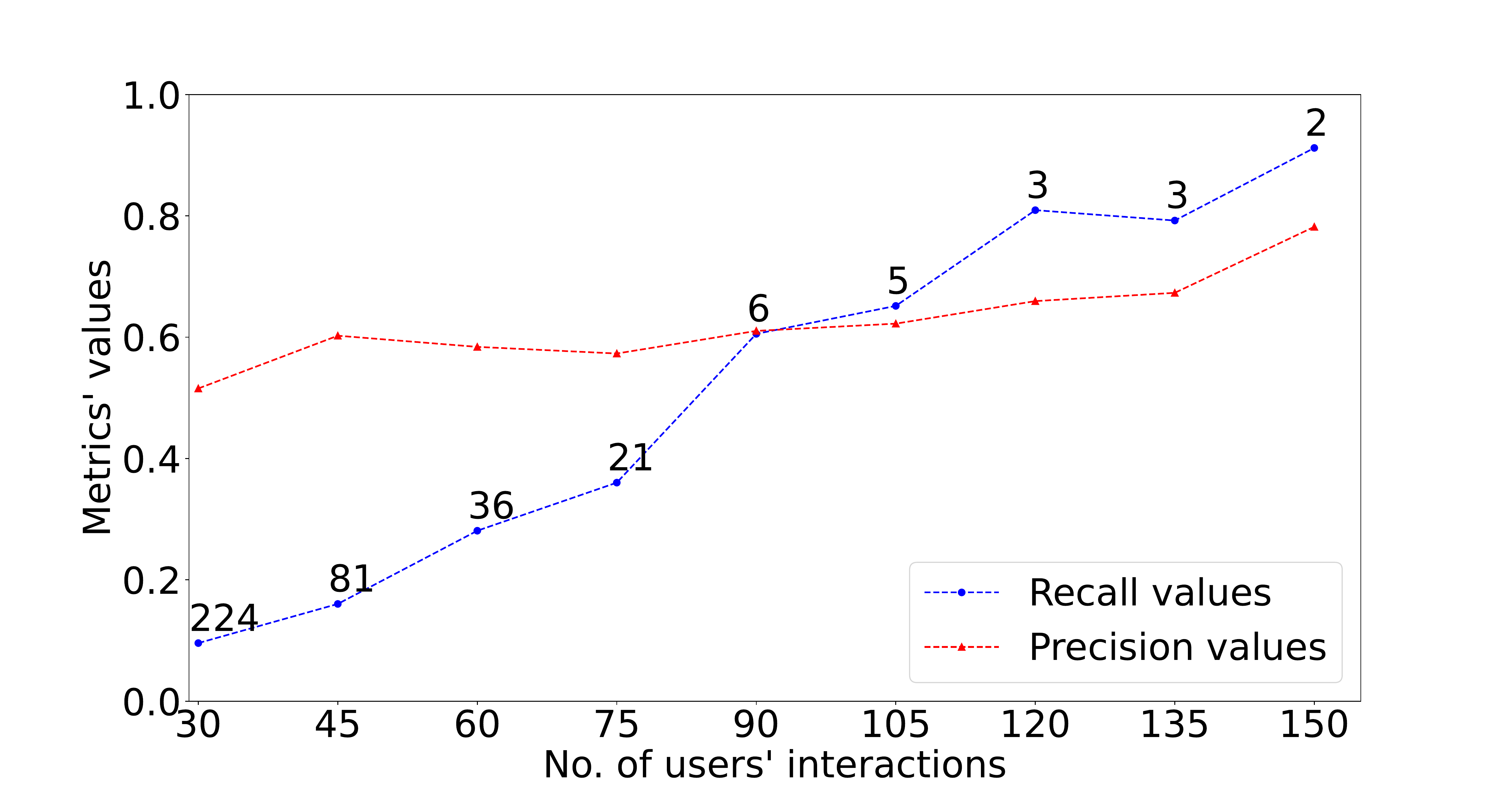}
  \end{subfigure}
  \caption{Evolution of PTER performance results (AUC-PR, AUC-ROC, recall and precision) by varying the ``\#active users" hyper-parameter. The annotations in the dots indicate this value. The Barcelona dataset was used for this experiment.}
  \label{fig:pter-overall-performance}
\end{figure}

PTER performance is directly affected by this trade-off, as we can see in Figure \ref{fig:pter-overall-performance}. As expected, PTER performs better when decreasing the number of active users, that is, when the matrix to learn is less sparse.

To adjust our model, we separated the pre-processed input reviews with their target output vectors into three different partitions. (1) The training partition (70\% of the dataset), (2) the validation partition (15\% of the dataset) and (3) the test partition (the remaining 15\% of the dataset). These partitions were randomly shuffled in every experiment. In order to tackle the sparsity issue, we took two decisions. First, we double the importance of the positive labels (representing the user contexts in Equation \ref{form:general_f_learn}) during the training stage. With this aim, we defined the ``1's labels weight" hyper-parameter, so that we modified the loss function, as we will see later. The second decision was to address the imbalance between positive and negative labels using the ML-ROS-20 random oversampling method \cite{oversampling_ml_ros} in the training partition. Basically, ML-ROS-20 computes a set of metrics of a MLD (Multi-Label Dataset) to check the initial label distribution. It determines if the dataset is unbalanced based on those metrics and therefore requires oversampling. If so, the minority labels constitute a ``bag" from which new samples are randomly drawn and added (repeated) in the training partition. This process increased the ratio of the minority labels\footnote{As a side effect, this oversampling method tends to balance the ratios of 1's throughout all the outputs labels (i.e., like an equalisation process).}. These two decisions combined, made the model more sensitive to positive detections and mitigated the interactions-per-user sparsity problem.

Since BERT weights are frozen, we only focus on the connections of the second block of our network during training. Our training objective was to minimise the BCE (Binary Cross Entropy) loss function, which is the suitable function for a multi-label classification task. We leveraged a customised version of this function based on the aforementioned ``1's labels weight" hyper-parameter, in order to weigh double the positive labels with respect to the negative labels, which mean absence of information about user contexts. During training, we followed an early stop method, monitoring the validation loss of this customised BCE function, using both a $\Delta$ (minimum change in the monitored metric between two consecutive epochs to qualify as an improvement) and a ``patience epochs" (number of epochs with no improvement after which training will be stopped) hyper-parameters. We used the Adam optimisation algorithm, following the original proposal of BERT \cite{bert}, as well as the recommended values for the batch size and learning rate in fine-tuning tasks\cite{bert}. We followed the naive approach of keeping the $N$ head tokens of the reviews \cite{text_guide_truncation_method, flair_framework}, on the assumption that the most meaningful information is written first. The ``input sequence length" hyper-parameter matches the maximum length constraint of BERT \cite{bert}. The ``\#active users" hyper-parameter was set accordingly with the performance curves in Figure \ref{fig:pter-overall-performance}. For the sake of comparison, we set the same value for every city. For the rest of hyper-parameters, values were set by trial and error accordingly with the validation data during the training stage. Table \ref{tab:pter-hyper-parameters} contains the values of the hyper-parameters employed to train the model. 

We enhanced the generalisation capabilities of the model using (1) a dropout layer, (2) L2 regularisation and (3) early stopping. Early stopping is a common strategy to guarantee the best fit of parameters, avoiding overfitting to the training data. For our proposed LTR supervised task, the early stop process made the executions stop in the range of 8-15 epochs, depending on the dataset.

\begin{table}[!htb]
\centering
\resizebox{\textwidth}{!}{%
\begin{tabular}{|c|c|c|c|c|c|c|c|c|c|c|}
\hline
\textbf{\begin{tabular}[c]{@{}c@{}}Batch\\ size\end{tabular}} & \textbf{\begin{tabular}[c]{@{}c@{}}Adam\\ learning \\ rate\end{tabular}} & \textbf{\begin{tabular}[c]{@{}c@{}}Review\\ policy\end{tabular}} & \textbf{\begin{tabular}[c]{@{}c@{}}Input\\ length\end{tabular}} & \textbf{\begin{tabular}[c]{@{}c@{}}\#Active \\ users\end{tabular}} & \textbf{\begin{tabular}[c]{@{}c@{}}ML-ROS\\ training \\ partition\\ increase\\ (\%)\end{tabular}} & \textbf{\begin{tabular}[c]{@{}c@{}}1's labels\\ weight in\\ BCE loss \\ function\end{tabular}} & \textbf{\begin{tabular}[c]{@{}c@{}}Dropout \\ rate\end{tabular}} & \textbf{\begin{tabular}[c]{@{}c@{}}Weight of \\ L2 reg term\end{tabular}} &
\textbf{\begin{tabular}[c]{@{}c@{}}Val \\ loss\\$\Delta$\end{tabular}} &
\textbf{\begin{tabular}[c]{@{}c@{}}Patience \\ epochs\end{tabular}}\\ \hline
16 & 3e-5 & \begin{tabular}[c]{@{}c@{}}Keep N \\ head \\tokens\end{tabular} & 512 & 100 & 20\% & 2 & 0.1 & 0.001 & 0.01 & 3 \\ \hline
\end{tabular}%
}
\caption{General and customised PTER model hyper-parameters.}
\label{tab:pter-hyper-parameters}
\end{table}

Assessing the usefulness and the validity of the explanations that PTER provides is a complicated issue due to the subjective nature of the concept of explanation. To assess our PTER model, we centred on the following 3 core ideas: (1) we designed an heuristic method to select our explainable reviews from the rankings, (2) we used the authorship criterion of the reviews to check the human suitability of explanations in a clustering process and (3) we defined a random adversary to make a comparison.

Given the $reviews^+(res_k) ~ \forall res_k \in R$ in the test and validation sets, the aim of our proposal was to obtain new dyads $(u_j, res_k, custom_{jk})$ where $custom_{jk}$ is the personalised explanation that was selected based on the rankings using the PTER model. In order to check the validity of our model, we guaranteed that we had a real positive review written by $u_j$ about $res_k$, so that we already had the reference dyad $(u_j, res_k, author_{jk})$. Then, we competed against a random adversarial dyad definition $(u_j, res_k, random_{jk})$.

In order to perform the evaluation, we filter each predicted ranking to preserve a top $N$ positive reviews (with $N$ a small number of the overall ranking size), which we identified as the ``predicted context" of each interaction $(u_j \xrightarrow{s} res_k)$. This $N$ evaluation parameter was tailored for every city dataset to keep only the reviews with the highest probabilities. The ``predicted context" helped us extracting meaningful text-based features, weighing the importance of every restaurant feature in terms of text and according to the output user.

To get the most meaningful text-based features of the ``predicted contexts", we considered the following filtering criteria for the reviews:

\begin{itemize}
    \item We discarded all non-alphanumeric characters from the reviews.
    \item We filtered English stopwords\footnote{The corpora that was used is available at  \url{https://www.nltk.org/nltk_data/}.}, which are high frequency words without relevant semantics (e.g., demonstratives, prepositions and so on).
    \item We followed POS tagging, selecting the singular nouns\footnote{The English off-the-shelf tagger, using the Penn Treebank tagset, was used. Available at \url{https://www.ling.upenn.edu/courses/Fall_2003/ling001/penn_treebank_pos.html}.} as the most meaningful English part of speech.
    \item We applied lemmatization, to get the root from word inflections \footnote{The lemmatization process leverages the WordNet's built-in morphy function.}.
\end{itemize}

After that, we encoded these ``predicted contexts" of reviews to a TF-IDF (Term Frequency - Inverse Document Frequency) vector space model representation, applying a tokenizer (fit to the whole vocabulary $V_{city}$ of a $city \in C$). TF (Term Frequency) weighs the occurrences of a word of $V_{city}$ in a review (or set of reviews, considering the ``predicted context"). IDF (Inverse Document Frequency) represents the specificity of the term (i.e., give more importance to the terms that appear only in few reviews). Using this TF-IDF measures, we mapped the ``predicted contexts" corresponding to each interaction $(u_j \xrightarrow{s} res_k)$ to a single cumulative vector. We deleted from these vector representations the top 20 words with the highest frequencies in the vocabulary of the reviews in each city. In this way, we discarded the most common, generic and meaningless words (e.g., food, restaurant, dish...) from the specific user contexts.

To select the personalised and explainable review from the ``predicted contexts" of each interaction $(u_j \xrightarrow{s} res_k)$, we followed this heuristic process:

\begin{enumerate}
    \item We took, from the ``predicted context" cumulative vector of each interaction $(u_j \xrightarrow{s} res_k)$, the positions with the highest weights (i.e., \textit{keywords}).
    \item We individually computed the frequency of each \textit{keyword} in all the reviews from the ``predicted contexts".
    \item We selected the personalised and explainable review from the ``predicted context" based on a score, computed for every review as the addition of the \textit{keywords} frequencies, penalised logarithmically with the ranking position.
    \item The review with the highest score within the ``predicted context", added to the interaction $(u_j \xrightarrow{s} res_k)$, explains the new dyad $(u_j, res_k, custom_{jk})$.
\end{enumerate}

We give more importance to the highest positions in the rankings, also considering the most important \textit{keywords} of the ``predicted context" for every interaction $(u_j \xrightarrow{s} res_k)$. It is worth mentioning that the selected review, $custom_{jk}$, can have a position different from the first in the rankings.

We present two examples of the reviews selected as the explanations, given a restaurant and two different users in the Barcelona dataset. In the first example, two users (``John W" and ``LaurentG752") are linked to the restaurant ``Restaurante\_Salamanca". Their keywords vary according to their interactions and therefore the selected reviews. Looking at the keywords from the user ``John W", they are mainly related to the seafood and the service of the restaurant, as the selected review does. On the other side, the keyword ``beach" of the user ``LaurentG752" changes the selected review, this time specifying that the restaurant is right next to the beach in the Mediterranean sea.

\begin{lstlisting}[language=bash, keywords={smooth}, basicstyle=\small\ttfamily, mathescape=true]
$\longrightarrow$ Pair ("John W", "Restaurante_Salamanca")
- Keywords from predicted context : ["paella", "seafood", "service"]
- Selected review at ranking #1: "When you see this restaurant do not be intimidated by how busy it looks. Service is very quick and friendly. Do not hesitate to ask them for advice regarding the food. I had the Lubina, which was recommended to me. The food is fresh and good. But whta won me over was the service. Eventhough the place is packed, the staff always make time for a little chat and offers service with a genuine smile. Their sangria is also delicious. And their calamari is very well prepared (no rubbery rubbish here). It is the perfect place for a layed back lunch or dinner."

$\longrightarrow$ Pair ("LaurentG752", "Restaurante_Salamanca")
- Keywords from predicted context : ["paella", "seafood", "beach"]
- Selected review at ranking #6: "The seafood and paella at this place are unreal. I asked many locals where the best paella was, they said Salamanca. They were so right. It was a bit pricey but you get what you pay for. I've never had seafood like that anywhere near the Mediterranean. And to top it off, they a great selection of Jamon. So so so good...and it's right next next to the beach."

\end{lstlisting}

In the second example, two users (``Paul B" and ``iamnotfatimbeyonce") are linked to the restaurant ``Teresa\_Carles". In this particular example, ``Paul B" is more interested in vegan food, so the selected review is tailored in that way. On the other hand, the user ``iamnotfatimbeyonce" prefers information regarding breakfasts (``excellent fresh juices").

\begin{lstlisting}[language=bash, keywords={smooth}, basicstyle=\small\ttfamily, mathescape=true]
$\longrightarrow$ Pair ("Paul B", "Teresa_Carles")
- Keywords from predicted context : ["teresa", "vegan", "food"]
- Selected review at ranking #4: "It was my second time to visit Teresa Carles and the food is just really tasty! I love the atmosphere and the fact that as a vegan you can try almost anything on the menu. Warmly recommend!"

$\longrightarrow$ Pair ("iamnotfatimbeyonce", "Teresa_Carles")
- Keywords from predicted context : ["food", "juice", "breakfast"]
- Selected review at ranking #23: "Breakfast beautifully presented and tastey. Excellent fresh juices."

\end{lstlisting}

\section{Experimental Results}
\label{sec:experimental_results}

We carried out the experimentation using the six datasets \cite{tripadvisor_datasets} described in Table \ref{tab:datasets_distribution_reviews_users}. In Section \ref{subsec:pter-performance}, we assessed the performance of the PTER model and the suitability of the ground truth assumption with a set of well-known classification metrics in the test partitions. In Section \ref{subsec:clustering-experimentation}, we checked the human suitability of selected explanations comparing them with a random selector. We have followed a clustering process considering the authorship criterion of reviews. For all these experiments, we have used the hyper-parameters in Table \ref{tab:pter-hyper-parameters}.

Moreover, we also have included a comparative study between data quantity and data quality in the datasets in Section \ref{subsec:comparative_study}. From this study we concluded that, in dyadic contexts,  more data does not mean better performance. We rely on the ``density" of the interactions per user. Finally, in Section \ref{subsec:extra-framework} we conducted a comparative study of PTER with other state-of-the-art baselines using the EXTRA benchmark datasets \cite{extra-framework}.

\subsection{PTER performance experimentation}
\label{subsec:pter-performance}

In Table \ref{tab:review_counts}, we provide the evolution of the number of reviews throughout the different stages of data preparation. We used the random ML-ROS-20 oversampling method in the training partition. In Table \ref{tab:test_results}, we present the metrics obtained with 5 experiment repetitions using the test partition. We retrained the second block of the network in every experiment, providing mean $\mu$ and standard deviation $\sigma$ considering random shuffling of all the partitions. 

\begin{table}[!htb]
\centering
\resizebox{0.8\textwidth}{!}{%
\begin{tabular}{|c|c|c|c|c|c|c|}
\hline
\textbf{Dataset} & \textbf{\begin{tabular}[c]{@{}c@{}}\#Reviews\\ (English)\end{tabular}} & \textbf{\begin{tabular}[c]{@{}c@{}}\#Reviews\\ (positive\\ reviews \\ {[}4-5{]})\end{tabular}} & \textbf{\begin{tabular}[c]{@{}c@{}}\#Reviews\\ (discard\\ empty\\ values\\ and nulls)\end{tabular}} & \textbf{\begin{tabular}[c]{@{}c@{}}\#Reviews\\ (discard\\ zero-valued\\ targets)\end{tabular}} & \textbf{\begin{tabular}[c]{@{}c@{}}\#Reviews\\ (train\\ ------\\ test/val)\end{tabular}} & \textbf{\begin{tabular}[c]{@{}c@{}}ML-ROS-20\\ Oversampling\\ (train \\ partition)\end{tabular}} \\ \hline
New Delhi & 148303 & 123095 & 119289 & 106419 & \begin{tabular}[c]{@{}c@{}}74493\\ ------\\ 15963\end{tabular} & 89406 \\ \hline
New York & 517604 & 424823 & 413669 & 400464 & \begin{tabular}[c]{@{}c@{}}280324\\ ------\\ 60070\end{tabular} & - \\ \hline
Madrid & 177353 & 145177 & 137705 & 102409 & \begin{tabular}[c]{@{}c@{}}71687\\ ------\\ 15361\end{tabular} & 86047 \\ \hline
London & 998939 & 833453 & 827592 & 813409 & \begin{tabular}[c]{@{}c@{}}569387\\ ------\\ 122011\end{tabular} & - \\ \hline
Barcelona & 417240 & 339385 & 334762 & 275783 & \begin{tabular}[c]{@{}c@{}}193048\\ ------\\ 41368\end{tabular} & 231667 \\ \hline
Paris & 510084 & 401589 & 398737 & 295228 & \begin{tabular}[c]{@{}c@{}}206660\\ ------\\ 44284\end{tabular} & 248010 \\ \hline
\end{tabular}%
}
\caption{\textit{\#Reviews}, considering filtering steps progressively. The ``discard zero-valued targets" step is related to the ``\#active users" hyper-parameter from Table \ref{tab:pter-hyper-parameters}. Test and validation partitions have the same amount of reviews. When the oversampling conditions of the ML-ROS method are not fulfilled, the process is omitted ``-".}
\label{tab:review_counts}
\end{table}

\begin{table}[!htb]
\centering
\resizebox{0.9\textwidth}{!}{%
\begin{tabular}{|c|c|c|c|c|c|c|c|}
\hline
\textbf{Dataset} & \textbf{\begin{tabular}[c]{@{}c@{}}AUC-PR \\($\mu \pm \sigma$)\end{tabular}} & \textbf{\begin{tabular}[c]{@{}c@{}}AUC-ROC \\($\mu \pm \sigma$)\end{tabular}} & \textbf{\begin{tabular}[c]{@{}c@{}}Precision \\($\mu \pm \sigma$)\end{tabular}} & \textbf{\begin{tabular}[c]{@{}c@{}}Recall \\(TPR) \\($\mu \pm \sigma$)\end{tabular}} &
\textbf{\begin{tabular}[c]{@{}c@{}}Specificity \\(TNR) \\($\mu \pm \sigma$)\end{tabular}} &
\textbf{\begin{tabular}[c]{@{}c@{}}Balanced \\ accuracy \\ (bACC) \\($\mu \pm \sigma$)\end{tabular}} & \textbf{\begin{tabular}[c]{@{}c@{}}F-measure \\($\mu \pm \sigma$)\end{tabular}} \\ \hline
Barcelona & $ 0.37 \pm 0.02$ & $0.80 \pm 0.01$  & $ 0.71 \pm 0.03$ & $0.20 \pm 0.02$ & $0.99 \pm 0.00$ & $0.57 \pm 0.00$ & $0.22 \pm 0.01$ \\ \hline
London & $0.39 \pm 0.01$ & $0.75 \pm 0.01$ & $0.57 \pm 0.01$ & $0.24 \pm 0.01$ & $0.97 \pm 0.00$ & $0.62 \pm 0.01$ & $0.37 \pm 0.01$  \\ \hline
Madrid & $0.35 \pm 0.04$ & $0.79 \pm 0.01$ & $0.79 \pm 0.03$ & $0.24 \pm 0.02$ & $0.99 \pm 0.00$ & $0.56 \pm 0.00$ & $0.18 \pm 0.01$ \\ \hline
New Delhi & $0.53 \pm 0.04$ & $0.84 \pm 0.01$ & $0.67 \pm 0.02$ & $0.42 \pm 0.02$ & $0.97 \pm 0.01$ & $0.63 \pm 0.01$ & $0.37 \pm 0.02$ \\ \hline
New York & $0.39 \pm 0.03$ & $0.78 \pm 0.02$ & $0.63 \pm 0.01$ & $0.23 \pm 0.04$ & $0.98 \pm 0.00$ & $0.63 \pm 0.01$ & $0.37 \pm 0.03$ \\ \hline
Paris & $0.39 \pm 0.04$ & $0.81 \pm 0.01$ & $0.78 \pm 0.04$ & $0.25 \pm 0.02$ & $0.99 \pm 0.00$ & $0.58 \pm 0.00$ & $0.25 \pm 0.01$ \\ \hline
\end{tabular}%
}
\caption{Mean $\mu$ and standard deviation $\sigma$ for the metrics computed using the test partition in 5 different runs.}
\label{tab:test_results}
\end{table}

In this context, a true positive (TP) means that the model predicted a positive label according to the ground truth, whereas a true negative (TN) is analogous for a negative label. Looking at the results in Table \ref{tab:test_results}, the low recall values with respect to very high specificity values may denote the sparsity and label imbalance problems in the datasets, stated in Subsection \ref{subsec:datasets}. We represented the balanced accuracy (bACC) as the fair metric for accuracy.

In terms of positive detections, the results show an overall good performance of the model in all datasets. Although we have low recall values indicating that the TP are less retrieved, the generally higher AUC-ROC and precision values indicate that the positives predicted by the PTER model are correct most of the time. This is an inkling about the consistency of our labelling proposal. 

New Delhi is the dataset with the best user context predictions, looking at the AUC-PR and AUC-ROC values. Despite the fact that in this dataset the precision is lower than in others, the recall is by far the best among them. New Delhi also has the best bACC value and the second higher F-measure value (considering $\sigma$ values). A good indicator about why New Delhi dataset outperforms the other datasets is the ``reviews/user ratio" from previous Table \ref{tab:datasets_distribution_reviews_users} in Section \ref{sec:materials_methods}. For this dataset, predicting user contexts is easier because its ``reviews/user ratio" is the highest (2.06) among the datasets. New Delhi interactions per user are more dense. On the contrary, Madrid is the dataset with the poorest performance. Again, referring to the ``reviews/user ratio" indicator from Table \ref{tab:datasets_distribution_reviews_users}, Madrid has the lowest value (1.64), hence the worst results if we consider the summary performance metrics bACC and F-measure.

\subsection{Clustering experimentation}
\label{subsec:clustering-experimentation}

\begin{figure}[!htb]
    \centering
    \includegraphics[width=1\textwidth]{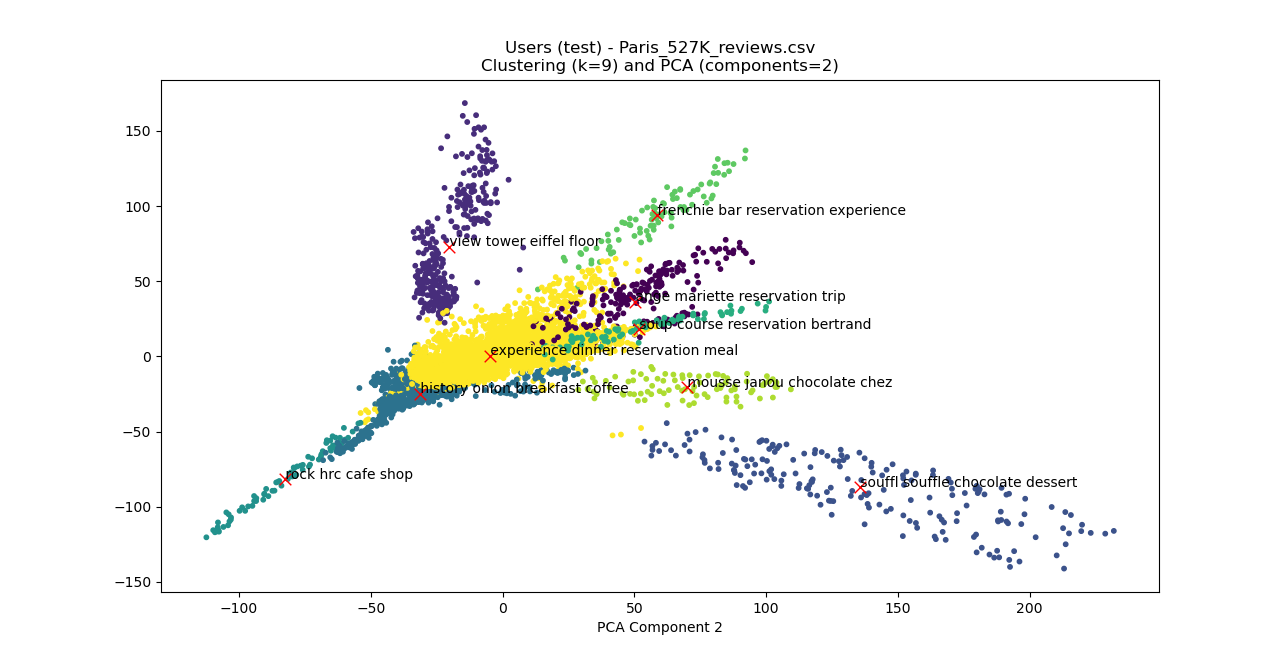}
    \caption{Example of the clustering process with $k=9$, using the TF-IDF ``predicted context" representations for the interactions $(u_j \xrightarrow{s} res_k)$, using the test partition of Paris dataset.}
    \label{fig:paris_k_9_clustering}
\end{figure}

To assess the usefulness and the validity of the explanations from the human perspective, we defined a clustering process where each data point corresponds to a TF-IDF cumulative vector from the ``predicted contexts" of an interaction $(u_j \xrightarrow{s} res_k)$ (see Section \ref{sec:proposed_method}). We used the data from the test partition to perform a k-means clustering. We followed the k-means++ algorithm for choosing the initial values and made the experimentation for $k \in \{3,5,7,9\}$. Figure \ref{fig:paris_k_9_clustering} depicts an example of this clustering using the Paris dataset. Each data point is depicted following a 2-D Principal Components Analysis (PCA). The centroids are labelled with the words with the highest weights in their TF-IDF vector representation. Therefore, these tags represent the main text-based features of the user contexts.

Once the clusters are obtained using the test partition, they are used to classify the $(u_j \xrightarrow{s} res_k)$ interactions of the validation partition. We classified three types of points: (1) the authorship points $(u_j, res_k, author_{jk})$ to be used as references, (2) the random points $(u_j, res_k, random_{jk})$ as the comparison adversary and (3) the personalised points $(u_j, res_k, custom_{jk})$ as the results of PTER to be assessed. In order to compare these types of points, we defined the Centroid Coincidence Ratio (CCR) metric as the number of times two points of different type are classified in the same centroid over a set of comparisons. 

To assess our model, we compared two different CCR values. We computed the CCR value of the authorship-random points and the CCR value of the authorship-personalised points. To do so, we followed these steps:

\begin{enumerate}
    \item First, we only preserved the interactions $(u_j \xrightarrow{s} res_k)$ from the validation partition where the user $u_j$ posted a review (or reviews) about the restaurant $res_k$. In that way, we guaranteed having the $(u_j, res_k, author_{jk})$ point (or points\footnote{If more than one review was available, we conducted more authorship point comparisons.}).
    \item For each new suggested interaction $(u_j \xrightarrow{s} res_k)$, we computed the personalised review following the heuristic method proposed in Section \ref{sec:proposed_method}. In this way, we generated our personalised and explainable version of the dyad $(u_j, res_k, custom_{jk})$.
    \item For each new suggested interaction $(u_j \xrightarrow{s} res_k)$, we computed a random review to generate the adversarial version of the dyad $(u_j, res_k, random_{jk})$. This review is randomly selected from the set of reviews available for $res_k$.
    \item We encoded the personalised, random and authorship versions into the TF-IDF vector space model. This process generates the three aforementioned types of points.
    \item We classified the three types of points in the clustering using euclidean distances to find their closest centroids.
    \item We computed CCR(A, R) for the authorship-random points and the CCR(A, P) for the authorship-personalised points.
    \item We computed the macro-averaged CCR value for the set of CCR(A, R) and the set of CCR(A, P).
    \item We compared the macro-averaged CCR(A, R) with the macro-averaged CCR(A, P).
\end{enumerate}

The results of this clustering evaluation using CCR are available in Table \ref{tab:ratio_coincidences_centroids}. As we can see, the majority of situations (Barcelona, New Delhi, New York, Paris and London) represent a clear improvement of our personalised approach with respect to the random adversary (looking at $\Delta(P-R)$ values for every $k \in \{3,5,7,9\}$). New Delhi dataset, which had the best results in PTER performance (see Table \ref{tab:test_results}), shows again the best results with the highest $\Delta(P-R)$ values (reaching a 18.5\% difference for $k = 9$). On the contrary, Madrid dataset, which had the worst results in PTER performance, shows the worst $\Delta(P-R)$ values. In fact, Madrid is the exception where our proposal does not work, being comparable to just randomly select a review and present it as the explanation to the user. We know that this is due to the poor quality of interactions defining the user contexts. As stated before, Madrid has the lowest value of the ``review/user ratio" (1.64) among the datasets, while New Delhi has the highest value (2.06). These metrics were introduced in Table \ref{tab:datasets_distribution_reviews_users} from Section \ref{sec:materials_methods}.

For the lowest value of $k = 3$, the tags that define clusters are more general and $\Delta(P-R)$ is lower, specially in London and New York datasets. This is because random points have more probabilities to match with authorship points in the same centroid (i.e., higher CCR(A, R) due to lower $\#centroids$). Apart from Barcelona, if we increase to $k = 5$, the difference $\Delta(P-R)$ widens. At this point, clusters are more specific and more appropriate for our personalisation proposal. If we continue increasing the value, $k \in \{7, 9\}$, the behaviour keeps the positive tendency for Barcelona, New Delhi and New York datasets. However, we start to see a negative tendency for Paris and London datasets, which could be a penalisation due to excessively targeted clusters.

In summary, we are comparing the PTER predictions with a random adversarial selector, using the authorship criterion to assess the human suitability of the explanations in a clustering process. \textit{We aim to see how near is our personalised and explainable approach to the original comments posted by the users, which are inherently the fairest positive explanation to an interaction} $(u_j \xrightarrow{s} res_k)$. If we are giving personalised explanations about why a restaurant $res_k$ is linked to a user $u_j$, the CCR(A, P) has to be higher with our personalised proposal. Roughly, we are representing what the user should have posted about that restaurant if he/she had visited that place before. Final results are improvable but acceptable considering that we followed the feature-based approach of the BERT weights (i.e., transfer learning task of a general purpose pre-trained transformer-encoder). It is important to mention that computational costs were significantly reduced during training time, since the 110M parameters of the transformer were frozen.

\begin{table}[!htb]
\centering
\resizebox{0.7\textwidth}{!}{%
\begin{tabular}{|c|c|c|c|c|c|c|}
\hline
\textbf{Dataset} & \textbf{Top N} & \textbf{\#A} & \textbf{k=3} & \textbf{k=5} & \textbf{k=7} & \textbf{k=9} \\ \hline
\begin{tabular}[c]{@{}c@{}}Barcelona CCR(A, P)\\ Barcelona CCR(A, R)\\ $\Delta(P-R)$\end{tabular} & 50 & 856 & \begin{tabular}[c]{@{}c@{}} \textbf{57.1\%} \\50.6\% \\ 6.5\% \end{tabular}  & \begin{tabular}[c]{@{}c@{}} \textbf{60.6\%} \\ 56.4\% \\4.2\% \end{tabular} & \begin{tabular}[c]{@{}c@{}} \textbf{46.3\%} \\ 41.2\% \\5.1\% \end{tabular} & \begin{tabular}[c]{@{}c@{}} \textbf{44.9\%} \\ 35.9\% \\ 9\%\end{tabular} \\ \hline
\begin{tabular}[c]{@{}c@{}}London CCR(A, P)\\ London CCR(A, R)\\ $\Delta(P-R)$\end{tabular} & 60 & 1491 & \begin{tabular}[c]{@{}c@{}} \textbf{90.8\%} \\ 90.7\% \\ 0.1\%\end{tabular} & \begin{tabular}[c]{@{}c@{}} \textbf{69.1\%} \\ 60.5\% \\8.6\% \end{tabular} & \begin{tabular}[c]{@{}c@{}} \textbf{69\%} \\ 61\% \\ 8\%\end{tabular} & \begin{tabular}[c]{@{}c@{}} \textbf{66.1\%} \\ 62.2\% \\ 3.9\% \end{tabular} \\ \hline
\begin{tabular}[c]{@{}c@{}}Madrid CCR(A, P)\\ Madrid CCR(A, R)\\ $\Delta(P-R)$\end{tabular} & 5 & 799 & \begin{tabular}[c]{@{}c@{}} 74\% \\ \textbf{75\%} \\-1\%\end{tabular} & \begin{tabular}[c]{@{}c@{}} 72\% \\ \textbf{72.2\%} \\ -0.2\% \end{tabular} & \begin{tabular}[c]{@{}c@{}} \textbf{66.3\%} \\ 66.2\% \\ 0.1\% \end{tabular}  & \begin{tabular}[c]{@{}c@{}} \textbf{47.1\%} \\46.4\% \\ 0.7\% \end{tabular}  \\ \hline
\begin{tabular}[c]{@{}c@{}}New Delhi CCR(A, P)\\ New Delhi CCR(A, R)\\ $\Delta(P-R)$\end{tabular} & 50 & 470 & \begin{tabular}[c]{@{}c@{}} \textbf{83.6\%} \\ 77.9\% \\ 5.7\% \end{tabular}  & \begin{tabular}[c]{@{}c@{}} \textbf{81.1\%} \\ 69.4\% \\11.7\%\end{tabular}  & \begin{tabular}[c]{@{}c@{}} \textbf{74.9\%} \\ 66.2\% \\ 8.7\%\end{tabular} & \begin{tabular}[c]{@{}c@{}} \textbf{67.4\%} \\ 48.9\% \\ 18.5\% \end{tabular}  \\ \hline
\begin{tabular}[c]{@{}c@{}}New York CCR(A, P)\\ New York CCR(A, R)\\ $\Delta(P-R)$\end{tabular} & 50 & 2498 & \begin{tabular}[c]{@{}c@{}} \textbf{80.6\%} \\ 80\% \\ 0.6\% \end{tabular} & \begin{tabular}[c]{@{}c@{}} \textbf{71.2\%} \\ 66.9\% \\4.3\% \end{tabular}  & \begin{tabular}[c]{@{}c@{}} \textbf{54.3\%} \\ 50.4\% \\ 3.9\% \end{tabular} & \begin{tabular}[c]{@{}c@{}} \textbf{52.6\%} \\ 47.3\% \\5.3\%\end{tabular} \\ \hline
\begin{tabular}[c]{@{}c@{}}Paris CCR(A, P)\\ Paris CCR(A, R)\\ $\Delta(P-R)$\end{tabular} & 50 & 672 & \begin{tabular}[c]{@{}c@{}} \textbf{69.2\%}\\ 65.8\% \\ 3.4\% \end{tabular} & \begin{tabular}[c]{@{}c@{}} \textbf{66.2\%} \\59.4\% \\6.8\% \end{tabular} & \begin{tabular}[c]{@{}c@{}} \textbf{61\%} \\ 56.1\% \\4.9\% \end{tabular} & \begin{tabular}[c]{@{}c@{}} \textbf{52.2\%} \\ 47.6\% \\4.6\% \end{tabular} \\ \hline
\end{tabular}%
}
\caption{Evaluation metric CCR(A, P) (Authorship, Personalised) and CCR(A, R) (Authorship, Random). $\Delta(P-R)$ measures the differences between both macro-averaged CCR values. ``Top N" corresponds to the evaluation parameter presented in Section \ref{sec:proposed_method}, defining the ``predicted contexts". \#A stands for the total number of authorship comparisons made, considering all the available reference reviews of the users. We varied the number of centroids $k \in \{3,5,7,9\}$. Bold text marks the best method for every $k$.}
\label{tab:ratio_coincidences_centroids}
\end{table}

\subsection{A comparative study: data quantity vs. data quality}
\label{subsec:comparative_study}

In dyadic data environments, more data does not mean better performance. In order to show the impact of other variables, we have measured the quality of input data as the ``density" of the interactions per user. To prove that data quality is more important than data quantity, we explored the sparsity of the interactions per user among the datasets. In order to make a fair comparison, we only compared among the datasets for which oversampling \cite{oversampling_ml_ros} was addressed. New Delhi was our comparison reference, as it is the dataset with the highest performance, although it is also the one with the lowest quantity of reviews.

\begin{figure}[!htb]
  \centering
  \begin{subfigure}[b]{0.7\textwidth}
    \includegraphics[width=\textwidth]{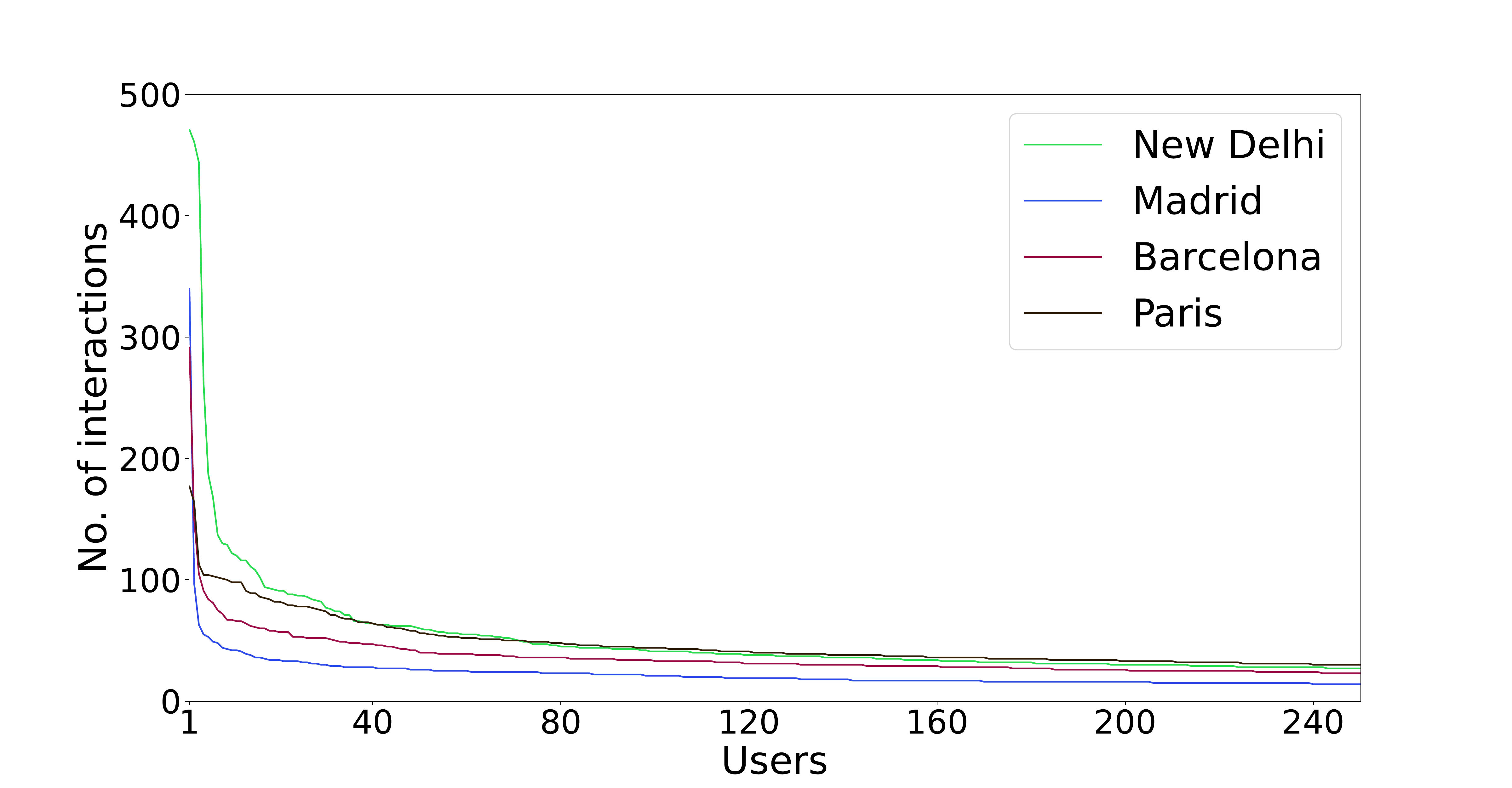}
  \end{subfigure}
  \par\bigskip
  \begin{subfigure}[b]{0.7\textwidth}
    \includegraphics[width=\textwidth]{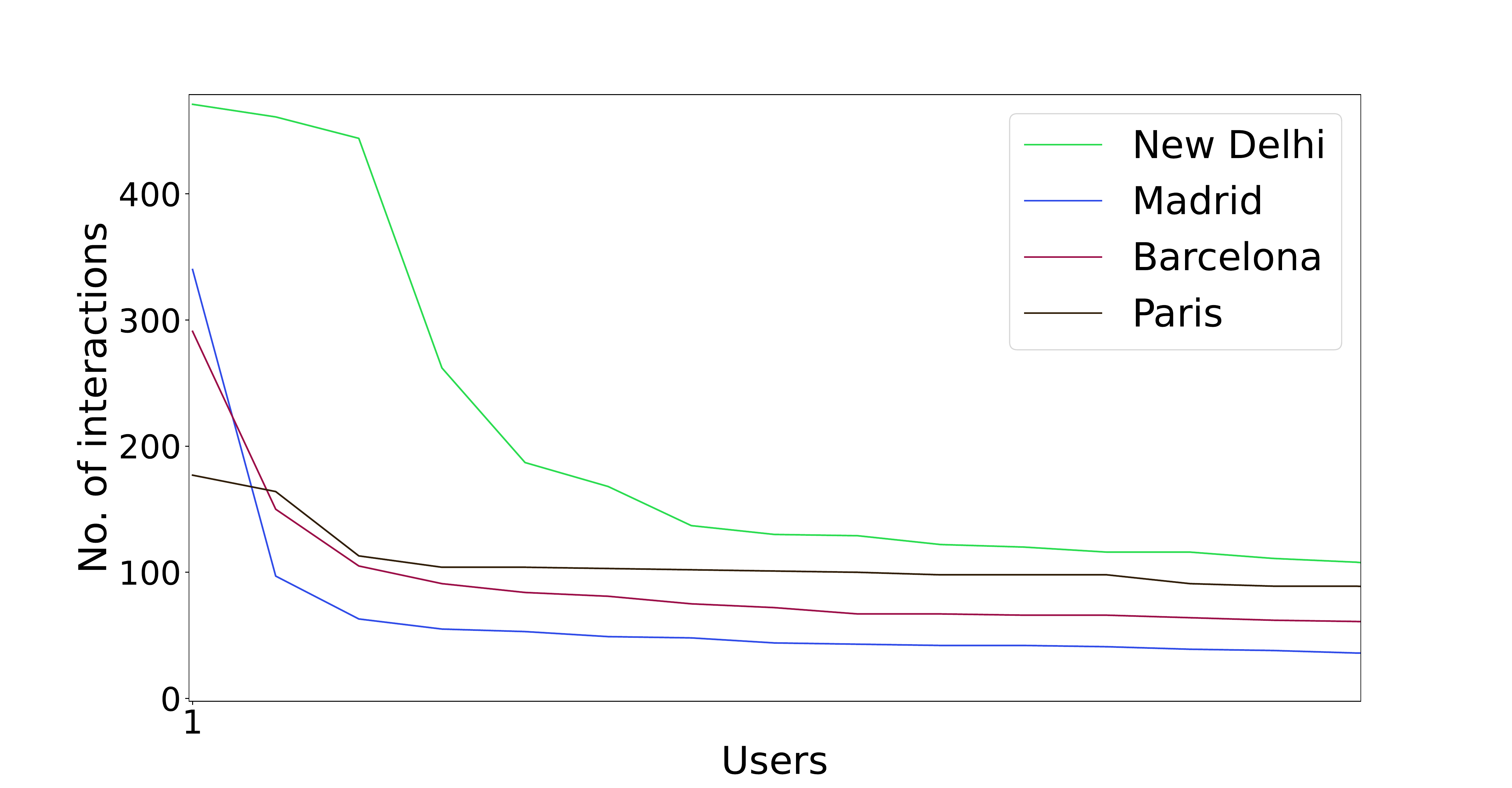}
  \end{subfigure}
  \caption{The long tail distribution and sparsity problem of user interactions, comparing New Delhi reference with Madrid, Barcelona and Paris datasets. Showing the most active users (top 250). Zooming in the head of the distribution in the second plot (top 15).}
  \label{fig:datasets_new_delhi_interactions}
\end{figure}

New Delhi and Madrid are the datasets with the best and the worst performance/evaluation metrics, respectively, even though these datasets have a very similar number of reviews (both are the smallest among the datasets, as depicted in Table \ref{tab:review_counts}). The significant differences in performance and evaluation rely on the data quality (i.e., the density of the interactions per user), as depicted in Figure \ref{fig:datasets_new_delhi_interactions}, where we present a comparison of their top users sorted by their number of interactions.

If we compare New Delhi with Barcelona, Figure \ref{fig:datasets_new_delhi_interactions} shows that a higher amount of reviews does not mean performance improvement. Barcelona dataset triples the amount of data, but the sparsity in the interactions per user is higher, thus the worse performance. The same applies to the comparison between New Delhi and Paris datasets, represented in the same Figure \ref{fig:datasets_new_delhi_interactions}. These differences are more appreciable when zooming in the region with the most active users.

The density of the user interactions is a clear quality indicator in Table \ref{tab:ratio_correlation}, where the correlation between the aforementioned ``review/user" ratio and the performance metrics is presented. The lower the ``review/user" ratio (i.e., the ``density"), the lower the performance metrics in every dataset.

\begin{table}[!htb]
\centering
\resizebox{0.7\textwidth}{!}{%
\begin{tabular}{|c|c|c|c|c|}
\hline
\textbf{Dataset} & \textbf{\begin{tabular}[c]{@{}c@{}}Review/user\\ ratio\end{tabular}} & \textbf{F-measure} & \textbf{AUC-PR} & \textbf{\begin{tabular}[c]{@{}c@{}}$\bm{\Delta(P-R)}$\\ $\bm{median ~(\%)}$\end{tabular}} \\ \hline
New Delhi & 2.06 & 0.37 & 0.53 & 10.2 \\ \hline
New York & 1.94 & 0.37 & 0.39 & 4.1 \\ \hline
London & 1.89 & 0.37 & 0.39 & 5.95 \\ \hline
Paris & 1.83 & 0.25 & 0.39 & 4.75 \\ \hline
Barcelona & 1.67 & 0.22 & 0.37 & 5.8 \\ \hline
Madrid & 1.64 & 0.18 & 0.35 & -0.05 \\ \hline
\end{tabular}%
}
\caption{Correspondence of the ``review/user" ratio w.r.t. the performance metrics F-measure, AUC-PR and the median of the evaluation metric $\Delta(P-R)$. Datasets are sorted by the ``review/user" ratio, from higher to lower. It is worth mentioning the correlation between this indicator and each of the metrics.}
\label{tab:ratio_correlation}
\end{table}

\subsection{EXTRA benchmark datasets}
\label{subsec:extra-framework}

To compare PTER in terms of explainability with respect to other baselines, we have used the EXTRA framework \cite{extra-framework} as it provides the results obtained by other systems, which are taken as baselines, on three benchmark datasets (TripAdvisor, Amazon, and Yelp). Our goal is twofold: 1) to compare the explainability, that can be measured using ranking-oriented metrics, and 2) to test the applicability of PTER in different dyadic contexts with new types of items (movies, series and hotels). It is worth mentioning that the sparsity issue of the user interactions is present also in this framework. The 5 baselines included in the framework for the dyad explainability task are:

\begin{itemize}
    \item RUCF (Revised User-based CF) and RICF (Revised Item-based CF) are collaborative filtering approaches providing scores relying on user and item neighbours respectively.
    \item CD (Canonical Decomposition) and PITF (Pairwise Interaction Tensor Factorisation) are tensor factorisation methods centred on the reconstruction of a (user, item, explanation) interaction cube.
    \item RAND is a random baseline similar to our ad hoc adversary proposed in Section \ref{subsec:clustering-experimentation}.
\end{itemize}

Our model PTER is a competitive model and beats RAND, which is non-personalised. PTER also outperforms the two collaborative filtering approaches (RUCF and RICF) in both Amazon and Yelp datasets, but it performs worse than RUCF in the TripAdvisor dataset. For the tensor factorisation methods, PTER outperforms CD but cannot surpass the strong baseline PITF, since it is a model specially designed to tackle the data sparsity issue \cite{pitf_baseline}. The rest of EXTRA baselines and PTER itself are more affected by this problem. We mitigated the data sparsity issue considering the ``active users" hyper-parameter and the ML-ROS oversampling method.

In summary, the metrics give our model a second position in the benchmark. It is worth mentioning that PTER explains and personalises (user, item) pairs only learning from the text features of the user interaction contexts, following the Formula \ref{form:general_f_learn}. We addressed the problem in a very different way, with a multi-label supervised LTR downstream task using the transformer architecture.

We conducted the experiments following the guidelines of the EXTRA framework. We preserved the negative interactions (i.e., with a negative score) and we omitted the filtering of zero-valued targets from the labelled input data, so that we worked with all the triplets (user, item, explanation) proposed in EXTRA \cite{extra-framework}. We adapted some of the original PTER hyper-parameters from Table \ref{tab:pter-hyper-parameters} due to the new datasets: Adam learning rate (2e-5), \#active users (500) and we deleted the positive weight in the BCE loss function. We also changed from the feature-based approach to the fine-tuning approach of BERT. Since now we are adapting the pre-trained BERT model to the downstream task, we no longer require the bi-LSTM network to fully process the last four hidden states \cite{tuning_vs_feature_based}. We trained the model end-to-end for 4 epochs without early stopping \cite{bert}. We split the datasets as described in the EXTRA framework, in the same partitions for the 5 runs providing the average performance with the standard deviation.

We evaluated our results with the ranking-oriented metrics presented in the framework (NDCG@10, precision@10, recall@10 and F1@10), following the proposed global-level explanation ranking task (i.e., the explanations in the test partition are globally shared for all items) \cite{extra-framework}. This ranking-based explainability assessment fits the best to our model, since we are proposing a LTR downstream task. We present both PTER and EXTRA baselines results in Table \ref{tab:pter_vs_extra_baselines}.

\begin{table}[!htb]
\centering
\resizebox{0.8\textwidth}{!}{%
\begin{tabular}{l|cccc}
\hline
 & NDCG@10 (\%) & Precision@10 (\%) & Recall@10 (\%) & F1@10 (\%) \\ \hline
 & \multicolumn{4}{c}{\textbf{Amazon Movies \& TV}} \\ \hline
CD & 0.001 $\pm$ 0.000 & 0.001 $\pm$ 0.000 & 0.007 $\pm$ 0.001 & 0.002 $\pm$ 0.000\\
RAND & 0.004 $\pm$ 0.000 & 0.004 $\pm$ 0.000 & 0.027 $\pm$ 0.003 & 0.006 $\pm$ 0.001 \\
RUCF & 0.341 $\pm$ 0.005 & 0.170 $\pm$ 0.003 & 1.455 $\pm$ 0.026 & 0.301 $\pm$ 0.005 \\
RICF & 0.417 $\pm$ 0.002 & 0.259 $\pm$ 0.003 & \underline{1.797 $\pm$ 0.019} & 0.433 $\pm$ 0.005 \\
PITF & \textbf{2.352 $\pm$ 0.025} & \textbf{1.824 $\pm$ 0.015} & \textbf{14.125 $\pm$ 0.157} & \textbf{3.149 $\pm$ 0.029} \\
PTER & \underline{1.672 $\pm$ 0.600} & \underline{0.268 $\pm$ 0.081} & 1.774 $\pm$ 0.611 & \underline{0.466 $\pm$ 0.144} \\ \hline
 & \multicolumn{4}{c}{\textbf{TripAdvisor}} \\ \hline
CD & 0.001 $\pm$ 0.000 & 0.001 $\pm$ 0.000 & 0.003 $\pm$ 0.001 & 0.001 $\pm$ 0.000 \\
RAND & 0.002 $\pm$ 0.000 & 0.002 $\pm$ 0.000 & 0.011 $\pm$ 0.001 & 0.004 $\pm$ 0.000 \\
RUCF\textbf{(*)} & 0.260 $\pm$ - & \underline{0.151 $\pm$ -} & \underline{0.779 $\pm$ -} & \underline{0.242 $\pm$ -} \\
RICF\textbf{(*)} & 0.031 $\pm$ - & 0.020 $\pm$ - & 0.087 $\pm$ - & 0.030 $\pm$ - \\
PITF & \textbf{1.239 $\pm$ 0.061} & \textbf{1.111 $\pm$ 0.037} & \textbf{5.851 $\pm$ 0.195} & \textbf{1.788 $\pm$ 0.059} \\
PTER & \underline{0.301 $\pm$ 0.019} & 0.085 $\pm$ 0.005 & 0.328 $\pm$ 0.019 & 0.135 $\pm$ 0.007 \\ \hline
 & \multicolumn{4}{c}{\textbf{Yelp}} \\ \hline
CD & 0.000 $\pm$ 0.000 & 0.000 $\pm$ 0.000 & 0.003 $\pm$ 0.001 & 0.001 $\pm$ 0.000 \\
RAND & 0.001 $\pm$ 0.000 & 0.001 $\pm$ 0.000 & 0.007 $\pm$ 0.000 & 0.002 $\pm$ 0.000 \\
RUCF\textbf{(*)} & 0.040 $\pm$ - & 0.020 $\pm$ - & 0.125 $\pm$ - & 0.033 $\pm$ - \\
RICF\textbf{(*)} & 0.037 $\pm$ - & 0.026 $\pm$ - & 0.137 $\pm$ - & 0.042 $\pm$ - \\
PITF & \textbf{0.712 $\pm$ 0.013} & \textbf{0.635 $\pm$ 0.008} & \textbf{4.172 $\pm$ 0.036} & \textbf{1.068 $\pm$ 0.013} \\
PTER & \underline{0.156 $\pm$ 0.021} & \underline{0.027 $\pm$ 0.004} & \underline{0.166 $\pm$ 0.023} & \underline{0.047 $\pm$ 0.007} \\ \hline
\end{tabular}%
}
\caption{Evaluation performance (\%) of PTER w.r.t. all the baselines from the EXTRA framework, providing the top-10 explanation rankings for the global-level explanation ranking task. We followed the experimentation details described in the EXTRA framework, with the same train and test partitions for the 5 experiment runs. We provide the average performance with the standard deviation. The best results are in bold and the second best results are underlined. \textbf{(*)} The benchmark paper only provides a single run for these experiments.}
\label{tab:pter_vs_extra_baselines}
\end{table}

\section{Conclusions}
\label{sec:conclusions}

PTER applicability is straightforward, since it is a ``presenter" for explaining, in a text-based and personalised way, new links between two different types of entities in dyadic environments. These links can be established by any agent (a recommender system, a paid promotion mechanism...). We proposed an approach in the specific context of the TripAdvisor platform, where entities correspond to the users and the restaurants. Our approach leverages the positive reviews posted about the restaurants presenting a LTR supervised task in several well-known cities, including Madrid, Barcelona, New York, New Delhi, London and Paris. Our main objective was to grasp user preferences based on their interactions, so that the personalised texts resemble the authorship and tastes of the users.

Using BERT as the word embedding generator is an advantage, thanks to its adaptability to a handful of NLP downstream tasks. The feature-based approach (i.e., freezing the transformer weights) significantly reduced the training time, following a transfer learning approach of the general purpose pre-trained model.

The PTER model was assessed in two different ways. Firstly, we designed an heuristic method to select our explainable reviews from the PTER rankings and compared them with a random adversarial selector in a clustering process. We used the authorship criterion of the reviews to prove the human suitability of the selected explanations. Secondly, we leveraged the EXTRA framework, following a ranking-oriented evaluation for explainability, comparing our model with other state-of-the-art baselines in dyadic data explanation. However, the scarcity of interactions per user account hindered the computation and the completeness of the evaluation metrics. Moreover, the lack of labelled data for the proposed downstream task made us define a ground truth. PTER gets better results when considering a subset of active users, since the sparsity issue in the labelling process is mitigated.

An interesting point for the future work is to distinguish between the two types of positive labelling in the BCE loss function. That is, give more importance to the authorship of the reviews in the ground truth definition. Moreover, we aim at extracting more datasets of other well-known cities in different languages, to prove the effectiveness of the model with a broader set of datasets. Our aim is to follow a multi-language PTER model approach, based on the multi-lingual transformer such as mBERT \cite{bert} or XLM-R \cite{XML-R}.

Moreover, the approach of personalised explanations using our TripAdvisor data was addressed exclusively with the positive reviews in the score range $[4-5]$. We aim at extending our proposal to the negative context of the restaurants. Negative reviews in the range $[1-3]$ provide valuable feedback to owners (i.e., clues to improve their services). We can leverage the negative interactions of the users and make analogous predictions, summarising the worst aspects that users will consider about a restaurant, trying to explain the ``negative" relationships in new dyads. Our evaluation framework can be extended to a negative context analogously.

Finally, an increasingly important notion related to both explainable and interpretable models is the need of falsifiable experiments with human participants. In \cite{falsifiable_interpretable_dnn}, the authors propose a high-level framework for best practices to generate robust and reliable interpretations, taking into account the falsifiability and the human verification. Other works present quantitative metrics to measure human explainability, including explanation goodness, explanation satisfaction, user understanding, user curiosity and user trust \cite{metrics_for_xai}. Crowd-sourcing with human participants has also been explored for explainable recommendation \cite{explanations_retrieve_text_based_narre, explanations_retrieve_text_based_der, nerar_model, text_templates_caesar}. However, this is harder to explore in our approach, since we rely on the public TripAdvisor platform and on the hundreds of thousands of public accounts, making unfeasible tackling the problem of direct human participation. We leveraged the authorship criterion as the implicit human assessment.

\section*{Acknowledgements}

This research has been financially supported in part by the Spanish Government research project PID2019-109238GB-C22, by the Xunta de Galicia under grant ED431G 2019/01, and by European Union ERDF Funds. Special recognition goes to the Spanish Ministerio de Universidades for the predoctoral FPU funds, grant number FPU19/01457. We would also like to thank CESGA (Centro de Supercomputación de Galicia) for giving us access to their computing resources.

\bibliography{pter-references}

\end{document}